# Public Policymaking for International Agricultural Trade using Association Rules and Ensemble Machine Learning

Feras A. Batarseh*[1] (corresponding author) *batarseh@vt.edu*

Munisamy Gopinath[2] *m.gopinath@uga.edu*

Anderson Monken[3, 4] *anderson.e.monken@frb.gov*

Zhengrong Gu[4] *zg120@georgetown.edu*

[1]Virginia Polytechnic Institute and State University (Virginia Tech); Bradley Department of Electrical and Computer Engineering; Arlington, VA, USA

[2]University of Georgia; Department of Agricultural and Applied Economics; Athens, GA, USA

[3]Federal Reserve Board of Governors; Washington, D.C., USA

[4]Georgetown University; Graduate School of Arts and Sciences; Washington, D.C., USA

**Abstract:** International economics has a long history of improving our understanding of factors causing trade, and the consequences of free flow of goods and services across countries. The recent shocks to the free-trade regime, especially trade disputes among major economies, as well as black swan events (such as trade wars and pandemics), raise the need for improved predictions to inform policy decisions. Artificial Intelligence (AI) methods are allowing economists to solve such prediction problems in new ways. In this manuscript, we present novel methods that predict and associate food and agricultural commodities traded internationally. Association Rules (AR) analysis has been deployed successfully for economic scenarios at the consumer or store level (such as for market basket analysis). In our work however; we present analysis of imports/exports associations and their effects on country-commodity trade flows. Moreover, Ensemble Machine Learning (EML) methods are developed to provide improved agricultural trade predictions, outlier events' implications, and quantitative pointers to policy makers.

**Keywords:** International Trade, Association Rules, Ensemble Machine Learning, Country-Commodity Transactions, Black Swan Events, Gravity Models.

## 1. Introduction

In recent years, many countries are concerned about rising trade deficits (value of exports less imports) and their implications on production, consumption, prices, employment, and wages. For instance, the United States' goods and services trade deficit with China was $345.6 billion in 2019 (USTR, 2020) Such large gaps are forcing countries to either exit trade agreements or enforce tariffs, (e.g. Brexit, U.S. tariffs on Chinese goods, and other de-globalization movements). Other outlier events could lead to additional uncertainties in global production

chains and irregularities in trade of essential goods (e.g. soybeans, steel, aluminum, and beef). While traditional economic models aim to be reliable predictors, we consider the possibility that AI methods (such as AR and EML) allow for better predictions to inform policy decisions. The 2020 U.S. National Artificial Intelligence Initiative Bill states: "Artificial intelligence is a tool that has the potential to change and possibly transform every sector of the United States economy" (S.1558 – AIIA, 2020). As AI methods are applied across different domains, they are often presented with a stigma of being irrelevant or a "black-box". Few studies have applied AI to economics; accordingly, some have called for increased attention to the role of AI tools (Mullainathan and Spiess, 2017). A study by the National Bureau of Economic Research (NBER) points to the potential of economists adopting ML more than any other empirical strategy (Athey and Imbens, 2017). While some success stories are recorded for deploying ML to international trade analysis, we present methods that elevate predictions' accuracy through *EML*, and evidence-based policy making through AR.

Open-government data provide the fuel to power the mentioned methods. Unlike traditional approaches, e.g. expert-judgment based linear models or time-series methods, EML and AR methods provide a range of data-driven and interpretable projections. Prior studies indicate the high relevance of AI methods for predicting a range of economic relationships with a greater accuracy than traditional approaches (Athey and Agrawal, 2009). With recent international trade policies garnering attention for limiting cross-border exchange of essential goods and with trade's critical effects on production, prices, employment, and wages (ERS, 2020), our motivation in this study is to test the following two-tier hypothesis: *(a) Trade transactions can point to country-commodity associations (i.e. AR) that are not revealed by traditional economic models. (b) Trade patterns are predicted on time and with higher accuracy* *if boosting models are using data from top world traders that are identified through clustering (i.e. EML).* Our contributions are summarized as follows: (1) developing a novel AR method for international trade, (2) providing trade predictions using EML that are superior to existing public sector predictions, (3) ML methods for the management of economic and trade policies during outlier events, (4) a review of the most important ML methods used in economics, (5) country-specific or commodity-specific trade predictions, and (6) an open source R dashboard that encompasses all the mentioned ML methods and visualizations.

Sections 1.1 and 1.2 present the motivation for AR and EML modeling. The next section (2) provides a review of traditional econometrics and presents the state of the art in applying AI methods to trade analysis. Section 3 introduces the methods (EML and AR) as they are applied to trade data; and section 4 presents all experimental results. Section 5 incorporates methods for validation, anomaly detection, outlier events and their manifestations on predictions. Section 5 also shows two use cases for policy making as well as other ideas for future work. Discussions, trade policy implications, and conclusions are presented in section 6.

1.1 Beer and Diapers for International Trade

The success of AR for online and brick and mortar commerce is undeniable. Amazon's "people who bought this item are also interested in…" is a recommendation engine that tells us what items have been bought together. Many other success stories exist: the famous beer and diapers association at grocery stores, Netflix's movie recommendations, Walmart's placing of bananas next to cereal boxes, and Target's predictions of pregnancy based on buying trends (MapR, 2020). Those commercial algorithms that use AR are referred to as MBA. In our work, we ask the question, can MBA through AR be applied to international trade? Can we look into trade of commodities between countries and build associations that could manifest trade associations such as: "*when the U.S. exports more Cereal to Mexico, Mexico will import less or no livestock and other animal products*". We look at trade flows and aim to illustrate how the waves of trade are influenced by an increase or decrease of commodity's flow between countries.

1.2 EML for Country-Commodity Trade

EML are meta-algorithms that combine several Machine Learning (ML) techniques into one model to decrease bias (such as through boosting) (Ke et al., 2017), and improve predictions (through using multiple methods). In this paper, we develop methods for trade predictions using boosting and k-means clustering. K-means clustering is a commonplace method for grouping items using non-conventional means. We cluster the countries of the world, and use the highest quality outputs of the k-means model to execute tailored predictions through boosting algorithms. Boosting is a class of ML methods based on the idea that a combination of simple classifiers (obtained by a *weak learner*) can perform better than any of the simple classifiers alone. Additionally, boosting is used to classify economic variables, and measure their influence on trade projections. A ranking of variables is presented per commodity to aid in understanding relationships between variables (for instance: GDP, Distance, and Population), and their effect on trade during conventional (white swan) and outlier (black swan) events.

**2. AI and Econometrics**

Econometrics (and model calibration, and other empirical methods used by economists) aim to elucidate mechanisms and identify causalities, whereas AI aims to find patterns in the data. The challenge with the latter for any kind of forward-looking or policy analysis is the famous "Lucas

Critique". Lucas (1976) argued that the economic parameters of traditional econometric models depended implicitly on expectations of the policy making process, and that they are unlikely to remain constant as events change and policymakers alter their behavior. In section 5, we present methods (for outlier detection and model's retraining) that aid policy makers in such situations, and allow them to amend their policies.

Section 2.1 presents an overview of AI methods applied to trade; and section 2.2 outlines traditional econometrics, and contrasts both paradigms.

## 2.1 Applying AI to Trade

Few studies have applied ML methods to economics. Storm et al. (2019) provided a comprehensive review of AI methods deployed to applied economics; especially their potential in informing policy decisions. In 2013, Gevel et al. published a book called "the Nexus between Artificial Intelligence and Economics". It was one of the first few works that introduced agent-based computational economics (Gevel et al. 2013). One year later, Feng et al. (2014) studied economic growth in the Chinese province of Zhejiang using a neural networks model. Their method however, is very limited in scope, and proved difficult to deploy across other provinces in China or other geographical entities in other countries. Abadie et al. (2010) developed a similar model, but applied it to the rising tobacco economy in California. In 2016, Milacic et al. (2016) expanded the scope, and developed a model for growth in GDP including its major components: agriculture, manufacturing, industry, and services. See also Kordanuli et al. (2016) for an application of neural networks for GDP predictions. Falat et al. (2015) developed a set of ML models for describing economic patterns, but did not offer predictions. In a recent study, Batarseh et al. (2019) presented boosting ML methods for trade predictions. In their study, many economic features were considered to identify which of them have the highest influence on trade predictions, and which ones could be controlled and tuned to change the forecasts. Different commodities had different rankings of economic variables, however population and GDP of both countries and tariffs had some of the highest impact on whether two countries would trade major commodities or not. AI methods are categorized into 3 areas: ML (including supervised and unsupervised), Deep Learning (DL), as well as Reinforcement Learning (RL). Table 1 presents a brief and relevant history of ML/DL/RL applications to international trade analysis.

Table 1: A brief history of AI methods for economics

| Year | AI Area | Short description | Reference |
|------|---------|------------------|-----------|
| 1999 | DL | Neural Networks in Economics and Quantitative Modeling | (Keilbach et al., 1999) |
| 2002 | ML | Approximate Factor Models | (Bai and Ng, 2002) |
| 2014 | ML | Predicting Winning and Losing Business with Tariffs | (Granell et al., 2014) |
| 2015 | RL | Fixing Energy Tariff Prices | (Serrano et al., 2015) |
| 2016 | ML | A latent variable model approach to work embeddings | (Arora et al., 2016) |
| 2016 | DL | Economic Growth Forecasting | (Mladenovi c et al., 2016) |
| 2016 | ML | Estimating Trade Policy Effects | (Piermartini et al., 2016) |
| 2016 | ML | Trade and Inequality | (Helpman et al., 2016) |
| 2016 | ML | Model for International Trade of Sawn Wood | (Nummelin et al., 2016) |
| 2018 | ML | Machine Learning in Agriculture | (Liakos et al., 2018) |
| 2018 | ML | Analysis of Bilateral Trade Flow | (Sun et al., 2018) |

| 2018 | DL | Neural Network Analysis of International Trade | (Wohl and Kennedy, 2018) |
|------|------|------|------|
| 2018 | ML | Forecasting Economic and Financial Time Series | (Namin and Namin, 2018) |
| 2019 | ML/DL | The Economics of Artificial Intelligence | (Agrawal et al., 2019) |
| 2019 | ML | The US-China Trade War | (Ciuriak, 2019) |
| 2019 | ML | Insurance Tariff Plans | Henckaerts et al., 2019) |
| 2019 | DL | Estimating China's Trade | (Dumor et al., 2019) |
| 2020 | DL/ML | Machine Learning in Gravity Models | (Munisamy et al., 2020) |

No method is found in literature that applied AR to country-commodity exports and imports (Serrano et al., 2015), and EML has not been extensively tested for international trade. The work presented in this paper addresses that gap.

2.2 Traditional Econometrics for Trade

A key objective of quantitative economic analyses is to uncover relationships – e.g. demand, supply, prices or trade – for use in making predictions or forecasts of future outcomes. However, when the current systems generates forecasts for decision making, they require a range of ad hoc, expert-driven or a combination of simple forecasting models supplemented by subject matter expertise to econometrics-based methods and mega-models, i.e. applied general equilibrium. Employing such approaches, many international institutions and government agencies project economic variables including trade flows to inform decisions in national and multilateral contexts (World Economic Outlook – International Monetary Fund, 2019; Trade in Goods and Services Forecast, Organization for Economic Cooperation and Development, 2019; World Trade Organization, 2019; U.S. Department of Agriculture, 2019). These predictions are highly valued by producer and consumer groups as well as policymakers in making decisions (Karali et al. 2019). However, some of these predictions based on a combination of simple linear models and expert judgment, have limitations (Isengildina-Massa et al. 2011). For instance, U.S. Department of Agriculture's (USDA) Outlook for Agricultural Trade has forecast accuracy that is below 35% (USDA, 2019), while the long-term U.S. Agricultural Projections to 2030 does not provide a measure assessing the quality of projections. Similarly, these predictions are also limited in their ability to quantify the contribution of underlying economic factors. For instance, Chapter 4 (World Economic Outlook – International Monetary Fund, 2019) notes that "the gravity model clearly distinguishes between the principal drivers of bilateral trade; these can be more difficult to disentangle in practice."

Mega-models (using applied general equilibrium techniques) have been a boon to trade policy research over the past 25 years (Kehoe et al., 2017). They have helped change the landscape of trade negotiations by arming developing and emerging economies with tools generally used by their developed counterparts (GTAP, 2019). However, as Kehoe et al. (2017) note, these models need additional theoretical development and improved measures of data employed for simulation of policy outcomes. These mega-models relying mostly on annual data are not nimble enough to respond to short- to medium-term informational needs. More importantly, policy uncertainty or uncertainty in general further limits both back-of-the envelope and mega-modeling approaches to inform decision-making. In the latter case, part of the challenge is the static nature of most trade models, which often conduct comparative static analysis of trade outcomes from deterministic trade policy changes. For example: Steinberg (2019) offers a theoretical model incorporating trade policy uncertainty. Using applied general equilibrium techniques, he finds that Brexit had a negative, but small effect on U.K.'s GDP. Other studies exploring the empirical effects of trade policy uncertainty or trade war include Amiti et al. (2019), Caldara et al. (2019) and Fajgelbaum et al. (2019).

Little guidance exists on theoretical modeling of trade policy uncertainty and its implications for producer and consumer behavior. As a result, ad hoc approaches to incorporating uncertainty can create specification bias in quantifying economic relationships and consequently, less precise outcomes on future agricultural trade patterns. The later, i.e. less precise forecasts, impacts producer and consumer decisions as well as government expenditures. These mega-models draw information from a variety of sources, e.g. elasticities, which can introduce additional specification errors or mismatch data distributions.

While grappling with trade policy and other uncertainties, the availability of big data and advances in AI software systems posed new challenges to conventional approaches (such as gravity models) to quantifying complex economic relationships (Varian 2014). The challenges include dealing with the sheer volume of data (evolving from spreadsheets and SQL databases towards Hadoop clusters and distributed data), the lengthy list of variables available to explain such relationships (and associated collinearity issues), and the need to move beyond static and linear models. AI has been offered as an alternative to address many of these challenges (Bajari et al. 2015; Mullainathan and Spiess 2017; Batarseh and Yang 2017; Athey and Imbens 2019). Several authors including Chief Economists of Google and Amazon have strongly advocated the use of big data and AI to uncover increasingly complex relationships even in an analysis as simple as fitting a supply or demand function. The economics community is catching on, but the speed of AI advances, i.e. new techniques emerge every month, can make an academic study stale by the time peer reviews are completed. Nonetheless, the academic community facing seismic shocks from advances in big data and AI has been called on to revisit time-tested theories and relationships (Mullainathan and Spiess 2017). In sum, traditional models – ad hoc, econometrics or mega-models – have been challenged both on modeling uncertainties, and providing accurate and on-time information for policy and decision making.

## 3. Methods

Data for this project are from the World Trade Organization (WTO) Bilateral Imports dataset, which comprises annual country to country trade data from 1996 to 2018. Data come from countries reporting imports from trading partners around the world, only countries that are part of the WTO report imports (the sample of countries is ~190). International trade data are organized by Harmonized Commodity Description and Coding System (HS Code System), and this study focused on the *chapter level* of the system: the 2-digit codes (HS-2). Data on the 96-chapter level trade products are downloaded from the WTO developer portal (https://apiportal.wto.org/). Using `Pandas` and `Numpy` libraries in Python, data are loaded into a *PostgreSQL* database for ease of analysis. We dig deeper (lower aggregation) for the EML models: data for EML are extracted from the General Agricultural Trade System (GATS) on the HS-4 level for prime agricultural commodities (years 1960-2018); from this web portal: https://apps.fas.usda.gov/gats/default.aspx.

Finally, International Monetary Fund (IMF) data are collected for GDP statistics (https://www.imf.org/external/datamapper/), and U.S. Department of Commerce data have been collected for example trade outliers in Livestock, and overall data and models' validation (presented in section 5). All cleaned datasets are shared in a public repository on Github (refer to the data availability statement).

## 3.1 AR for Trade

Mining frequent commodities that trade in association with each other has a direct implication on decision making. AR is a popular method for discovering hidden relations between variables in big datasets. Piatetsky and Shapiro (1991) describe analyzing *strong* rules discovered in datasets. Based on the notion of strong rules, Agrawal et al. (1993) introduced the problem of mining association rules from transaction data.

The idea of AR is as follows: Let $X = \{x_1, x_2, ..., x_n\}$ be a set of n commodities. Let $T = \{t_1, t_2, ..., t_n\}$ be a set of transactions (where $Tt$ is the total number of transactions). Each transaction in T has a unique ID and contains a subset of the commodities (X) that are traded. A rule is defined as an implication of the form $X_a \Rightarrow X_b$ where $X_a, X_b \subseteq X$ and $X_a \cap X_b = \emptyset$.

The sets of commodities $X_a$ is called antecedents (left-hand-side or LHS) and the set of commodities $X_b$ is called consequents (right-hand-side or RHS) of the rule.

Besides antecedent-consequent rules, the quality of the associations is measured through the following three metrics:

$$Support = \frac{X_a + X_b}{Tt} \quad (1)$$

$$Confidence = \frac{X_a + X_b}{X_a} \quad (2)$$

$$Lift = \frac{(X_a + X_b)/ X_a}{(X_b/Tt)} \quad (3)$$

Support indicates that for example 67% of customers purchased beer and diapers together. Confidence is that 90% of the customers who bought beer also bought diapers (confidence is the best indicator of AR). While lift represents the 28% increase in expectation that someone will buy diapers, when we know that they bought beer (i.e. lift is the conditional probability).

In our study, AR mining is performed using the `arules` library in R. Data are pulled and processed from PostgreSQL. Transactions in the data represent each country-country pair's trade for a given year. The goods in the data are the 96 commodity code trade dummies, which are boolean values depending on whether trade occurred for a specific country-country pair for a given year. Apriori association rules are collected with a minimum support of 0.35 and a maximum number of antecedents set to 3. Results are visualized in `plotly` and exported using R. The top 4 million+ rules are pulled out of the models, and migrated into a structured relational SQL database (called: `AR-Trade`). The rules are pulled for U.S. trade for top trade countries in Asia (China, Korea, and Japan); as well as the top trade countries in Europe (UK, Spain, France, and Germany). Relational SQL tables include the following columns: Lhs (antecedent), Rhs (consequent), Lhs name, Rhs name, Support, Confidence, Lift, Count, Country_O, Country_D.

AR results are plotted using an `R-Shiny` dashboard, as well as R plots using the `arulesViz` library such as through this script:

```
isS4(AR-Trade)
AR-Trade@lhs
plot(AR-Trade)
```

Results are recorded and analyzed. The next section presents EML methods (clustering and boosting), and how they are deployed in experimental setup.

## 3.2 EML for Trade

Python libraries are used to deploy three boosting trees (`GBoost`, `XGBoost`, and `LightGBM`) (Ke et al. 2017). Boosting trains weak learners sequentially, and in every cycle, each trying to correct its predecessor. Our boosting models offer an alternative to econometric approaches, which are seldom cross-validated.

K-means is used to cluster countries (to identify big world traders), different *k* values are used to improve the EML process, the best clustering model is chosen as input for the boosting algorithms. The k-means model is executed using Euclidean distances, where distance from p to q: *d (p, q) = Sqrt ((q1-p1)² + (q2-p2)² + ... + (qn-pn)²)*  (4)

Countries in certain clusters can be better indicators to international trade flows, for example, the trade values of France have higher influence on the predictions quality than the trade values of Fiji. Accordingly, in this EML process, we identify the best *k* for clustering using silhouette coefficient and elbow diagram methods.

Silhouette coefficient = max *s(k)*. Where *s(k)* represents the mean *s(i)* over all data points in the trade dataset for a specific number of clusters k. The k-means model is developed for *k=2* to *k=20*, "powerful" country clusters are then used to create better trade predictions/boosting. We aim to find the most appropriate Sum of Square Errors (SSE). For each k, a score is computed for SSE via the formula:

$$SSE = \sum_{n=1}^{n}(t_i^2) - \left.\sum_{n=1}^{n}(t_i)^2\right/ n \ (5)$$

SSE tends to decrease towards zero as k increases and so SSE is equal to 0 *when k* is equal to the number of data points in the trade dataset. Therefore, the goal is to choose a small value of *k* that has a low SSE. The result for all rows (n) and columns (t) is evaluated using an elbow diagram that indicates the most appropriate number of clusters. After clustering is performed, predictions are evaluated within the context of the clusters. Moreover, a "flag" validation and anomaly detection engine is deployed (with application to livestock) to allocate outliers and inform commodity-specific policy making; outcomes are collected and presented.

All data, code, and scripts for the experiment are available on GitHub (refer to the data availability statement).

## 4. Experimental Results

The models' results are illustrated through a R-Shiny dashboard, Tableau dashboard, other text tables, and R plots.

4.1 Descriptive Maps and AR Results

After the data are wrangled, few descriptive dashboards are developed, for instance, a heat map showing the history of trade is illustrated in Figure 1a. The rise of China is obvious in the figure, as well as how big traders seem to dominate international trade. Figure 1b illustrates a comprehensive international trade map that shows the flow of trade between all countries of the world. The highlighted example shows the mentioned Asian and European countries of choice, as well as the U.S. Additionally, the trade map presents the evolution of trade for every country; in Figure 1b, the UK is selected and trade values from 2010 to 2018 are illustrated in a bar chart. The Tableau dashboards presented in this manuscript are shared on Tableau public (a public online repository) and available for anyone for further exploration.

Certain countries trade more with each other due to treaties, distance, GDP, and other economic variables. Results from the associations (selected top confidence sums), for all countries are presented in Table 2 (showing agricultural and non-agricultural associations). An extended version of the table, with *top 100* rules is added to Appendix 1. Table 3 presents the top *multiple antecedent* rules by HS code, along with their confidence, support, and lift.

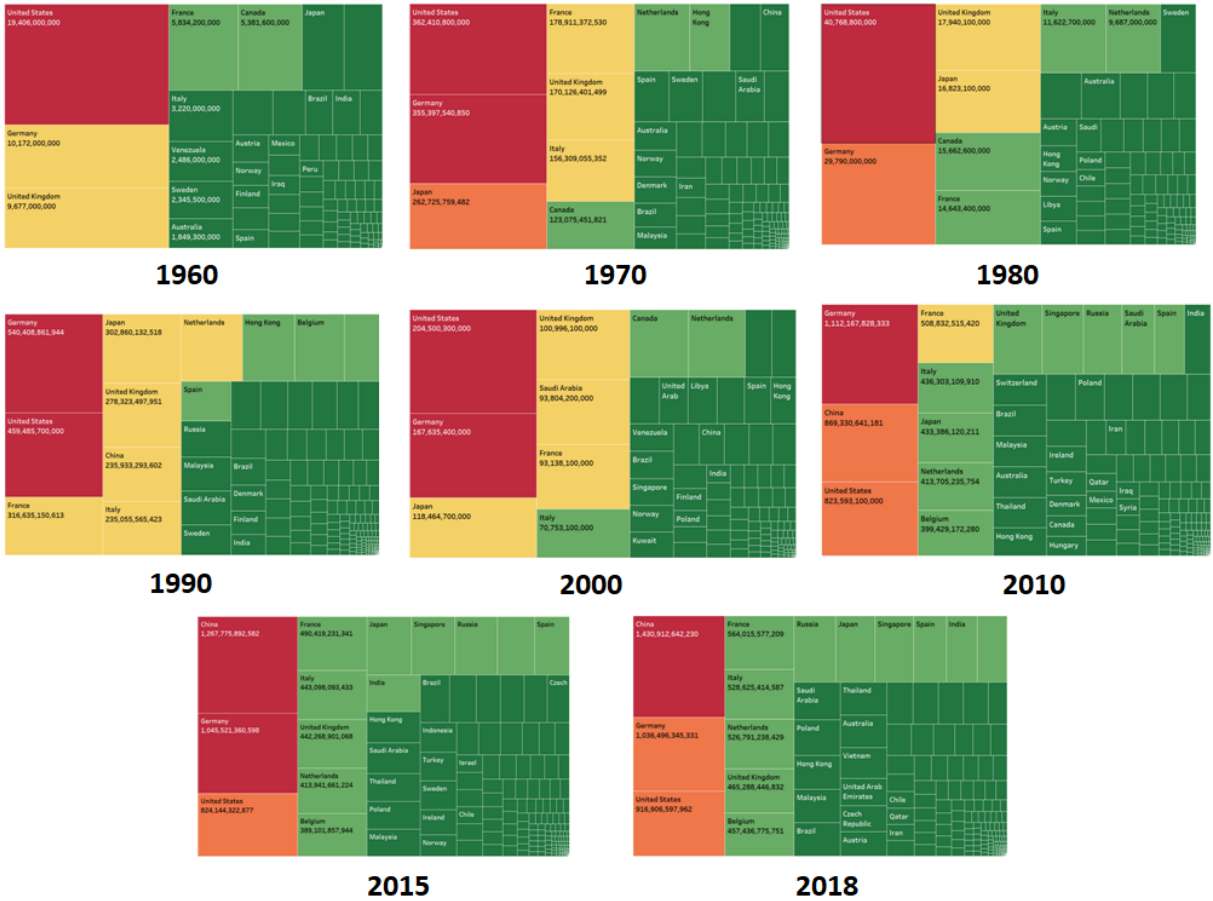

Figure 1(a): Tableau dashboard - a history of trade (China moving towards the lead)

Trade Map

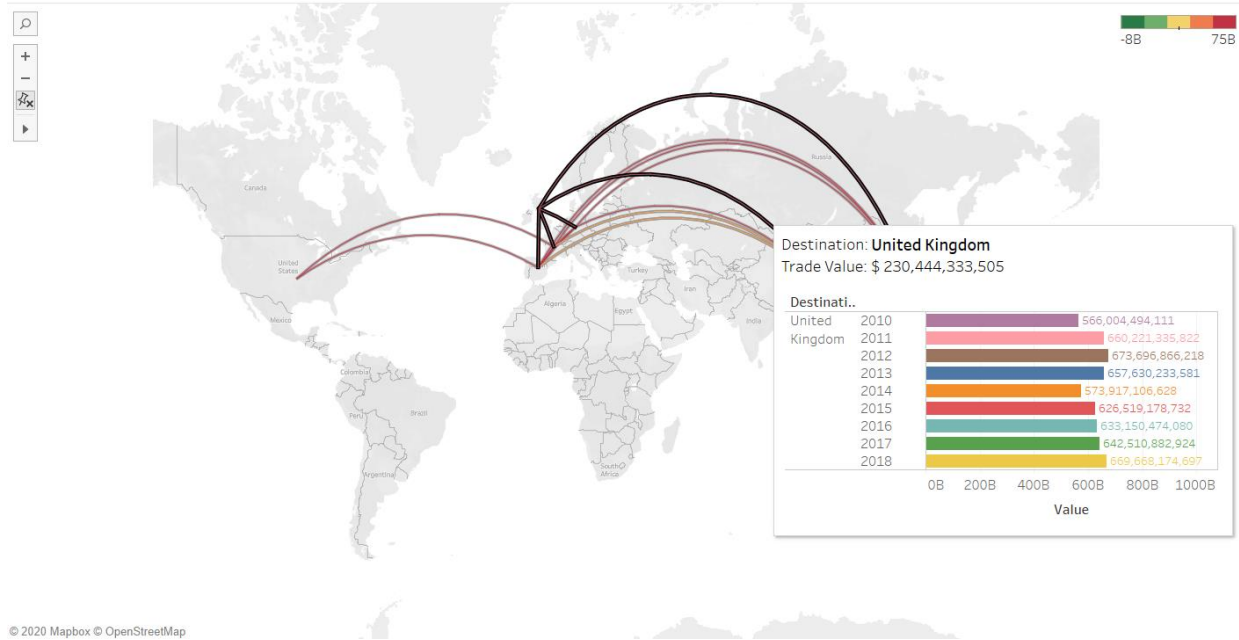

Figure 1(b): Tableau dashboard - international trade flows of country pairs

Table 2: Selected associations: agricultural and non-agricultural (5 rules each)

| Antecedent | Consequent | Sum of Confidence | Ag/Non-Ag |
|---|---|---|---|
| Cereals | Animal or vegetable fats and oils and their cleavage products; prepared edible fats; animal or vegetable waxes | 946 | Ag |
| Oil seeds and oleaginous fruits; miscellaneous grains, seeds and fruit; industrial or medicinal plants; straw and fodder | Cereals | 935 | Ag |
| Oil seeds and oleaginous fruits; miscellaneous grains, seeds and fruit; industrial or medicinal plants; straw and fodder | Clocks and watches and parts thereof | 924 | Ag |
| Animal or vegetable fats and oils and their cleavage products; prepared edible fats; animal or vegetable waxes | Cotton | 891 | Ag |
| Cocoa and cocoa preparations | Manmade filaments | 858 | Non-Ag |
| Preparations of vegetables, fruit, nuts or other parts of plants | Sugars and sugar confectionery | 847 | Ag |
| Preparations of cereals, flour, starch or milk; pastry cooks' products | Ores, slag and ash | 840.03 | Non-Ag |
| Beverages, spirits and | Explosives; pyrotechnic | 814 | Non-Ag |

| | | | | | |
|---|---|---|---|---|---|
| vinegar | products; matches; pyrophoric alloys; certain combustible preparations | | | | |
| Beverages, spirits and vinegar | Fertilizers | 814 | Non-Ag | | |
| Cocoa and cocoa preparations | Lead and articles thereof | 780 | Non-Ag | | |

Table 3: Top 20 multiple HS-code antecedent associations

| Antecedent {HS codes} | Consequent {HS codes} | *Support* | *Confidence* | *Lift* | Count |
|---|---|---|---|---|---|
| {11.0,21.0,74.0} | {19.0} | 0.405114 | 1 | 1.636564 | 301 |
| {11.0,33.0,69.0} | {19.0} | 0.401077 | 1 | 1.636564 | 298 |
| {11.0,21.0,65.0} | {19.0} | 0.432032 | 1 | 1.636564 | 321 |
| {11.0,21.0,40.0} | {19.0} | 0.430686 | 1 | 1.636564 | 320 |
| {11.0,21.0,73.0} | {19.0} | 0.440108 | 1 | 1.636564 | 327 |
| {28.0,38.0,65.0} | {76.0} | 0.405114 | 1 | 1.507099 | 301 |
| {28.0,65.0,82.0} | {76.0} | 0.411844 | 1 | 1.507099 | 306 |
| {28.0,65.0,70.0} | {76.0} | 0.411844 | 1 | 1.507099 | 306 |
| {28.0,65.0,83.0} | {76.0} | 0.419919 | 1 | 1.507099 | 312 |
| {28.0,40.0,65.0} | {76.0} | 0.430686 | 1 | 1.507099 | 320 |
| {39.0,76.0,88.0} | {40.0} | 0.403769 | 1 | 1.483034 | 300 |
| {76.0,88.0,94.0} | {40.0} | 0.401077 | 1 | 1.483034 | 298 |
| {30.0,73.0,82.0} | {40.0} | 0.437416 | 1 | 1.483034 | 325 |
| {22.0,54.0,65.0} | {33.0} | 0.401077 | 1 | 1.462598 | 298 |
| {16.0,19.0,34.0} | {33.0} | 0.405114 | 1 | 1.462598 | 301 |
| {32.0,70.0,96.0} | {33.0} | 0.402423 | 1 | 1.462598 | 299 |
| {21.0,32.0,34.0} | {33.0} | 0.403769 | 1 | 1.462598 | 300 |
| {32.0,34.0,65.0} | {33.0} | 0.402423 | 1 | 1.462598 | 299 |
| {32.0,34.0,64.0} | {33.0} | 0.405114 | 1 | 1.462598 | 301 |

| {22.0,29.0,32.0} | {33.0} | 0.401077 | 1 | 1.462598 | 298 |
| --- | --- | --- | --- | --- | --- |

Figure 2 is another part of the Tableau dashboard that shows the top trade countries, and allows the user to browse for associations of commodity-country pairs. Every "dot" is a rule, and the graph consists of the resulted 4 million+ rules. In the figure, the rule: (Carpets → Starches and Glues for USA → China) is selected. Figure 3 illustrates the top rules' and their affinity towards low support and high confidence.

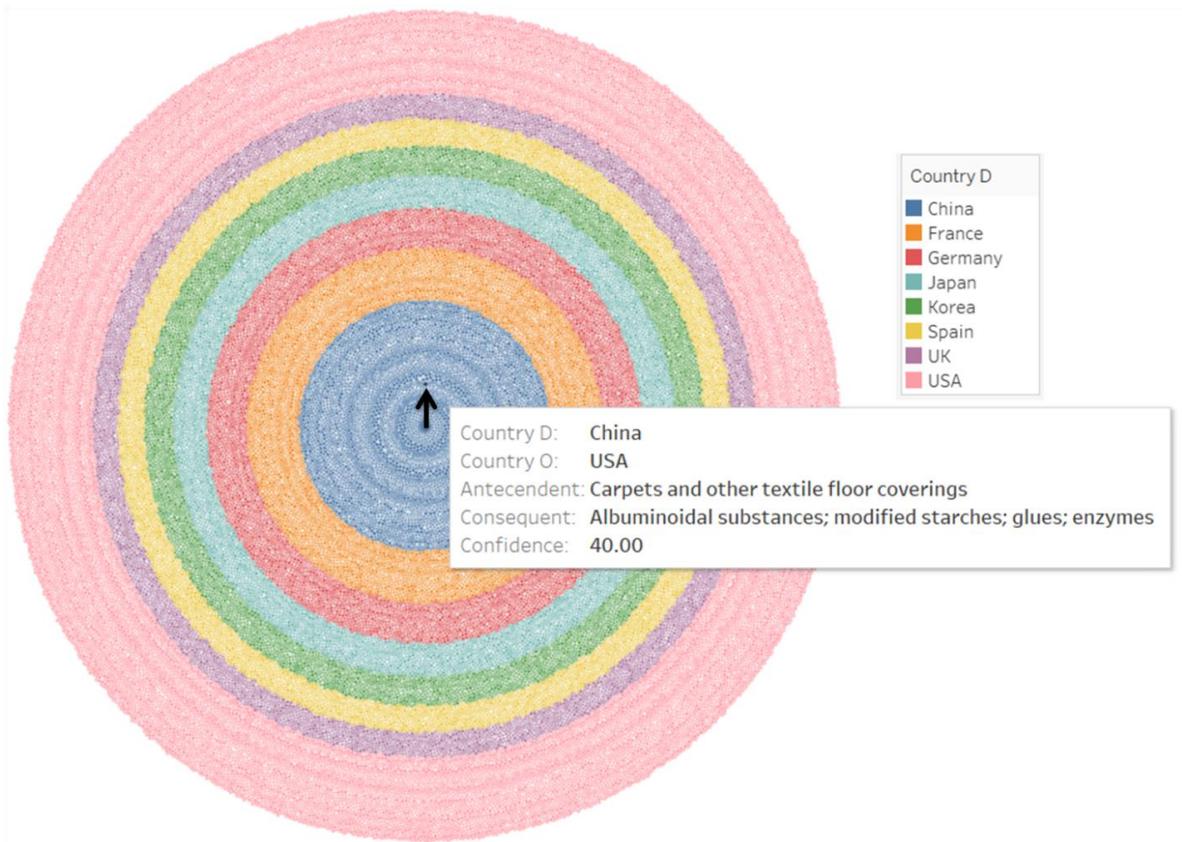

Figure 2: The 4 million+ Association Rules of Trade

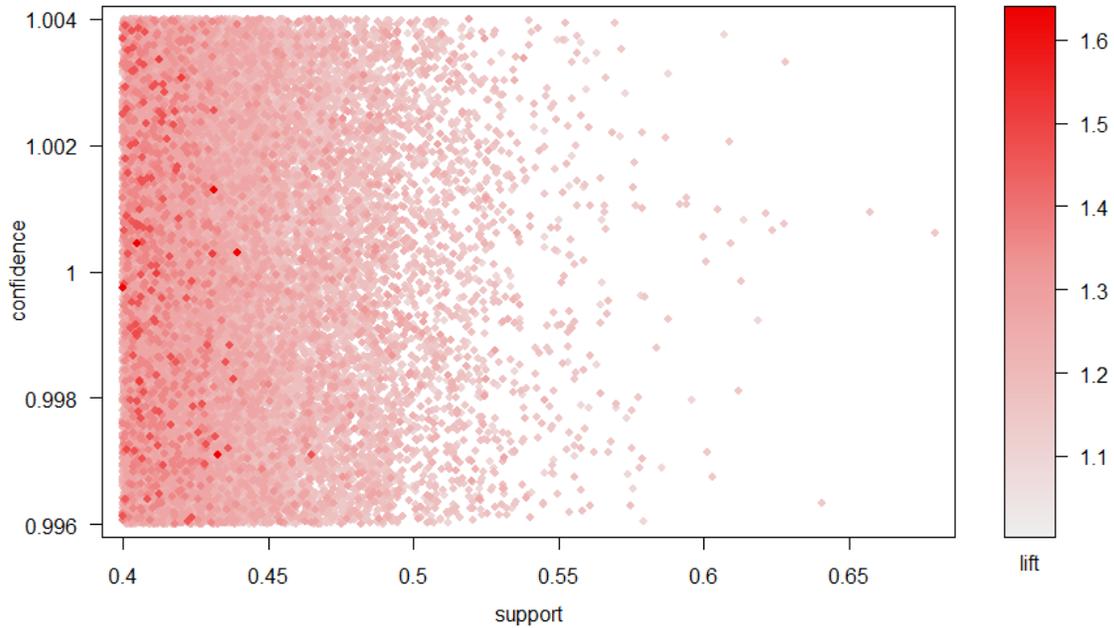

Figure 3: Support and Confidence scores of Trade AR

4.2 Ensemble Machine Learning Results

Table 4 illustrates how countries are belonging to different clusters based on the value of *k*. The table shows example 35 countries, the complete table of all countries is in Appendix 2. The presented elbow and silhouette measures (Figure 4) dictate that k=6 is the best model (the most appropriate SSE).

Table 4: Clustering of countries/regions for trade trends

| Country | k=2 | k=3 | k=4 | k=5 | k=6 | k=7 | k=8 | k=9 | k=10 |
|---------|-----|-----|-----|-----|-----|-----|-----|-----|------|
| **Argentina** | 1 | 2 | 2 | 2 | 2 | 5 | 1 | 3 | 1 |
| **Australia** | 1 | 2 | 1 | 1 | 5 | 4 | 4 | 1 | 3 |
| **Brazil** | 1 | 2 | 1 | 1 | 5 | 4 | 4 | 1 | 3 |
| **Cameroon** | 0 | 1 | 0 | 4 | 3 | 1 | 0 | 2 | 9 |
| **Canada** | 1 | 2 | 1 | 1 | 5 | 4 | 4 | 1 | 3 |
| **Chad** | 0 | 1 | 3 | 0 | 4 | 1 | 7 | 4 | 2 |
| **Chile** | 1 | 2 | 2 | 2 | 2 | 5 | 1 | 3 | 1 |
| **China** | 1 | 2 | 1 | 1 | 1 | 3 | 2 | 7 | 7 |
| **Cuba** | 0 | 0 | 0 | 3 | 0 | 0 | 5 | 6 | 8 |
| **Egypt** | 1 | 2 | 2 | 2 | 2 | 5 | 1 | 3 | 1 |
| **El Salvador** | 0 | 0 | 0 | 3 | 0 | 0 | 5 | 6 | 8 |
| **European Union** | 1 | 2 | 1 | 1 | 1 | 3 | 2 | 7 | 7 |
| **India** | 1 | 2 | 1 | 1 | 5 | 4 | 4 | 1 | 3 |
| **Indonesia** | 1 | 2 | 1 | 1 | 5 | 4 | 4 | 1 | 3 |
| **Israel** | 1 | 2 | 2 | 2 | 2 | 5 | 1 | 3 | 1 |
| **Jamaica** | 0 | 0 | 0 | 3 | 0 | 2 | 6 | 0 | 0 |
| **Japan** | 1 | 2 | 1 | 1 | 1 | 3 | 2 | 7 | 7 |
| **Malaysia** | 1 | 2 | 1 | 1 | 5 | 4 | 4 | 1 | 3 |
| **Maldives** | 0 | 1 | 3 | 0 | 4 | 1 | 7 | 4 | 2 |
| **Mali** | 0 | 1 | 3 | 4 | 3 | 1 | 7 | 4 | 4 |
| **Mauritania** | 0 | 1 | 3 | 4 | 3 | 1 | 7 | 4 | 4 |
| **Mauritius** | 0 | 0 | 0 | 4 | 0 | 2 | 6 | 8 | 0 |
| **Mexico** | 1 | 2 | 1 | 1 | 5 | 4 | 4 | 1 | 3 |
| **New Zealand** | 1 | 0 | 2 | 2 | 2 | 5 | 1 | 3 | 6 |
| **Niger** | 0 | 1 | 3 | 0 | 3 | 1 | 7 | 4 | 4 |
| **Nigeria** | 1 | 0 | 2 | 2 | 2 | 5 | 1 | 3 | 1 |
| **Panama** | 0 | 0 | 0 | 3 | 0 | 0 | 5 | 0 | 0 |
| **Saudi Arabia** | 1 | 2 | 1 | 1 | 5 | 4 | 4 | 1 | 3 |
| **Senegal** | 0 | 1 | 0 | 4 | 3 | 2 | 0 | 2 | 9 |
| **Singapore** | 1 | 2 | 1 | 1 | 5 | 4 | 4 | 1 | 3 |
| **South Africa** | 1 | 2 | 2 | 2 | 2 | 5 | 1 | 3 | 1 |
| **Switzerland** | 1 | 2 | 1 | 1 | 5 | 4 | 4 | 1 | 3 |
| **Ukraine** | 1 | 0 | 2 | 2 | 2 | 5 | 1 | 3 | 1 |
| **United States of America** | 1 | 2 | 1 | 1 | 1 | 3 | 2 | 7 | 7 |
| **Zambia** | 0 | 0 | 0 | 4 | 0 | 2 | 0 | 2 | 9 |

Figure 5 shows the R-Shinny app that is developed to present the 6-clusters model (as well as the associations and correlations), and how countries relate to each other in it.

The 6-clusters model illustrates the countries in cluster 1 are the biggest players (USA, Japan, China, EU…etc). That result is then used to predict the future of trade for year 2020 and beyond. For example, results for trade predictions through the XGBoost Model scored predictions' quality = **69%**, and through LightGBM scored a quality of **88%** (in contrast, GBoost scored the lowest of the three approaches). Data from cluster 1 countries are used to train the models, but scoring is applied to all countries. Parameter tuning for boosting models include: Number of leaves, Maximum Depth of the tree, Learning Rate, and Feature fraction.

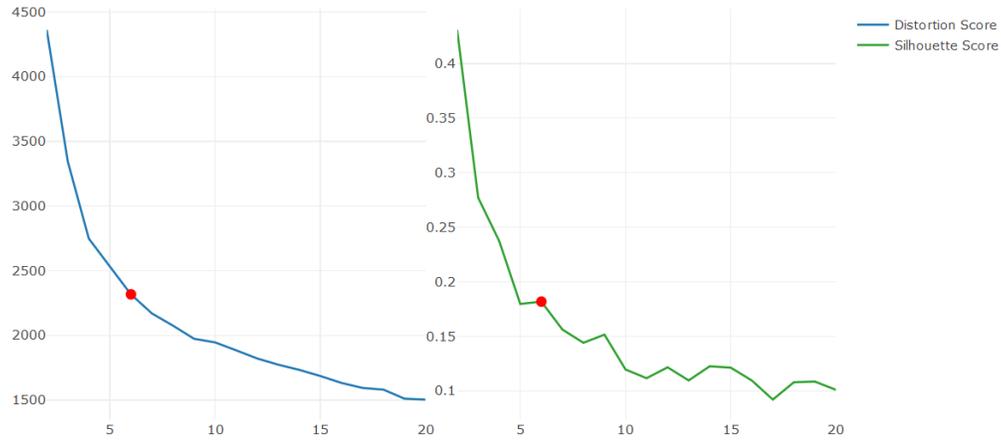

Figure 4: Silhouette and Elbow Diagrams

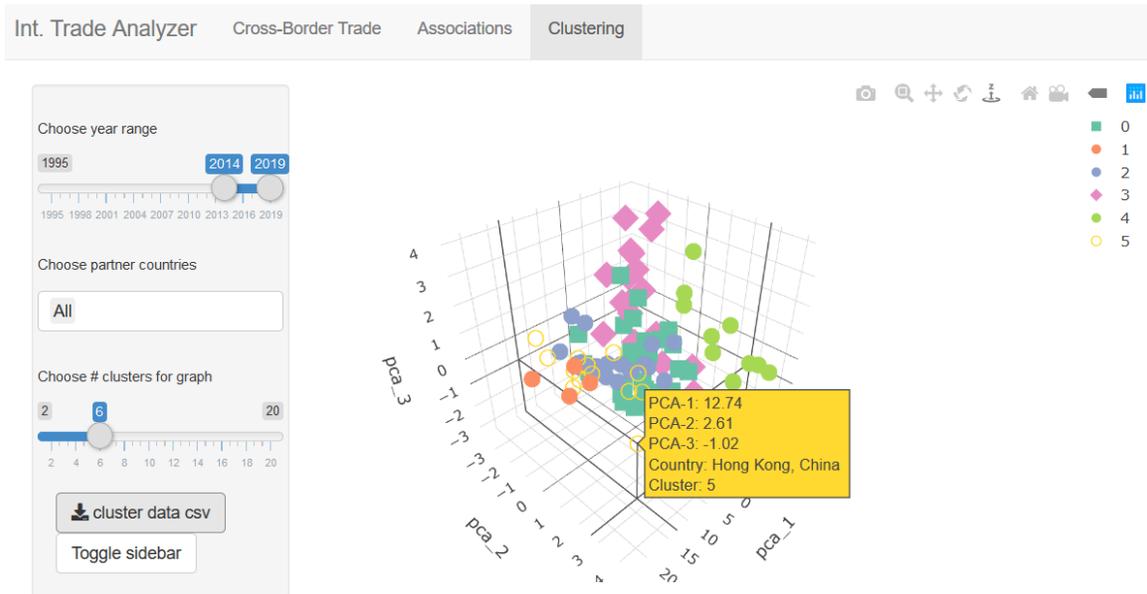

Figure 5: R-Shiny app for EML

Small learning rates are optimal (0.01), with large tree depths. Additionally, to speed up training and avoid over-fitting, feature fraction is set to 0.6; that is, selecting 60% of the features before training each tree. Early stopping round is set to 500; that allowed the model to train until the *validation score* stops improving. Maximum tree depth is set to 8.

Those settings led to the best output through LightGBM. *Sugar* for instance had an $R^2$ score (prediction quality) of 0.73, 0.88 for *Beef*, and 0.66 for *Corn* (three essential commodities). These initial results confirm the applicability of EML methods to projecting trade patterns and also point to accuracy gains over traditional approaches. However, after using only countries from cluster 1: the models yielded even better results. Figure 6 illustrates the boosting models quality (actual vs. predicted) before deployment of k-means outputs (i.e. using cluster 1), and after (i.e. with and without EML).

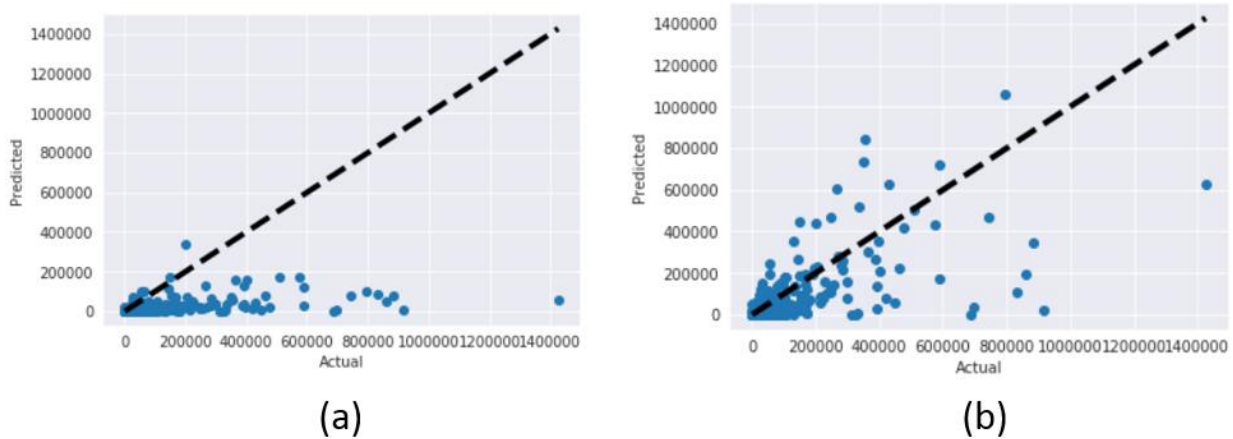

Figure 6: (a) Models quality without EML vs. (b) Models quality with EML

Moreover, EML methods presented in this manuscript allowed for the extractions of the best economic variables that would affect trade of specific commodities. *Distance* had the highest effect (i.e. the U.S.is better off trading beef to Canada and Mexico, its two closest neighbors). While Australia, being an island, has to focus its policies for beef exports on GDP measures, and the population of the importer. Feature Importance for all economic variables are (namely: split, gain.): Distance: 1469, 6.38. GDP of Exporter: 1431, 6.22. Year: 993, 4.318. Population of Exporter: 882, 3.83. Population of Importer: 847, 3.68. Currency of Importer: 801, 3.48.

The next section (5) introduces a use case for contrasting two commodities (milk and corn) by economic variables, and what that means for decision making. Over 2006-2017, the USDA forecast accuracy (Ag outlook reports) was less than 35% (USDA ERS, 2018). Their accuracy improves to 92% *only* after having actual data for three-quarters of the forecasted year. Models presented in this paper offer forecast accuracy in the 69% – 88% range (even before data for three quarters are available).

Albeit models presented offer higher accuracy, but what happens when those data patterns are majorly disrupted by an outlier event such as trade wars, embargoes, or pandemics? How are economic variables such as tariffs, production, and prices influencing policy during black swan events? The next section presents methods to evaluate such shifts in trade patterns and use AI methods to provide pointers to policy alternatives.

## 5.  Outlier Events and AI Methods

During the Covid-19 pandemic, the International Monetary Fund (IMF) updated their forecasts for the global economy. The IMF warned of soaring debt levels, as well as country's economies contracting (8% contraction for the US, 9.1% for Brazil, as well as 10.5% for Mexico). Moreover, the IMF estimates an overall contraction of 4.9% in global GDP in 2020 (IMF, 2020). Figure 7 illustrates IMF's Global GDP projections; the figure illustrates how Covid-19 lead to a major drop in 2020's country-wise GDP. Such numbers provide a general overview of the global economy's health; however, it is not clear how GDP rates will affect trade patterns around the world, as well as production, prices, and many other economic variables.

Isolation forests, distance-based, and density-based approaches can be generally used for outlier detection within a learning environment, however, for explainability and clarity purposes, in our presented boosting models, we present commodity-specific variable correlations to support policy making during outlier events. For example, some commodities infer that the GDP of the importer is the main influencer (via classification measures) of the size of trade between both countries. Other commodities exchange depends on distance or population and so on. Hence, by looking at commodity-specific changes (within different contexts), implications on policies that shape production, pricing, or consumption of that commodity could be altered in a more acute manner.

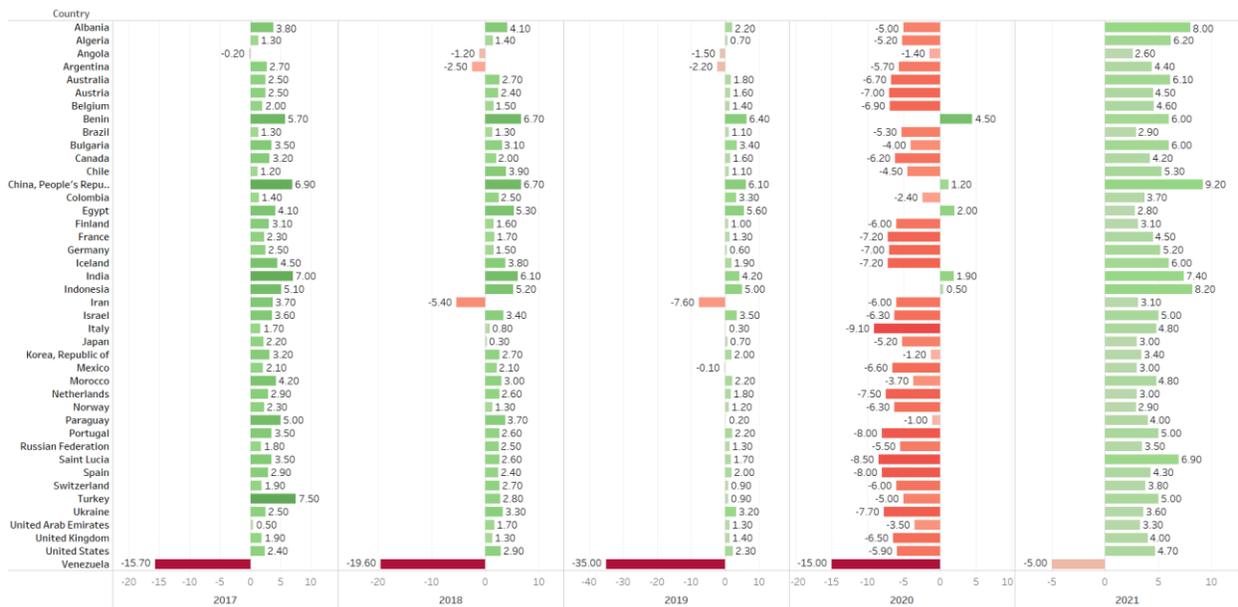

Figure 7: IMF Global GDP Trends (2017-2021): Negative GDP for most countries in year 2020

## 5.1 Data during Black Swan Events

In the aftermath of the Covid-19 pandemic, the Department of Justice (DOJ) subpoenaed the four largest meat packers in the US (Tyson, Cargill, JBS, National Beef). The DOJ looked into allegations of price manipulation as prices paid to ranchers declined while retail meat prices rose. The "Big Four" meat packers account for more than 70% of beef processing in the U.S. (Dorsett, 2020). Outlier events affect economic variables, and so during such black swan events, predictions and forecasts ought to be modified. During outlier events, historical data become much less dependable. Timeliness of data is a critical practical challenge as well. To estimate the current and future macroeconomic effects of Covid-19's uncertainties, we need measures that are available in real time, or nearly so. This kind of analysis can aid in avoiding a serious disruption to economic activity (Fan et al., 2020). During such events, analysis is performed in uncharted waters. Issues arise such as the need to use daily if not hourly data (but not monthly data) for pattern recognition and predictions. Additionally, decisions become timelier and need to be executed in a quick manner using real time analysis and on-demand analytics (Barro et al., 2020) (Eichenbaum et al., 2020).

Accordingly, as established prior, commodities flow and trade are influenced differently by different economic variables. During outlier times, traditional models fail to re-learn in real time to provide an updated prediction based on hundreds of variables. The box below presents a milk and corn use case to shed a light on determinants of trade patterns of these commodities using boosting.

Usecase#1:
While all industries have been seriously affected by the Covid-19 pandemic, food and agriculture have been among the hardest hit segments of the U.S. economy. The primary reason lies in the composition of household food expenditures. The impacts of the pandemic appear to vary by commodity based on two critical issues: perishability and labor use. Perishables like milk are among the hardest hit. Specialty crops such as corn also depend on labor for growing and harvesting – both types of commodities are affected during outlier events. Figure 8 illustrates feature importance (using classification tree splits) for milk trade and compares that to corn.
We consider commodity-specific distress during outlier events, for instance, on the biggest milk exporters: New Zealand, Germany and the U.S. If we know that GDP rates (as the IMF data showed) have sunk for those countries, and also, production is reduced due to closures and shutdowns; while other variables such as language, distance, and trade agreements stay the same, we are able to retrain the model with updated values and produce *new* real–time predictions that are suitable for the context at hand.

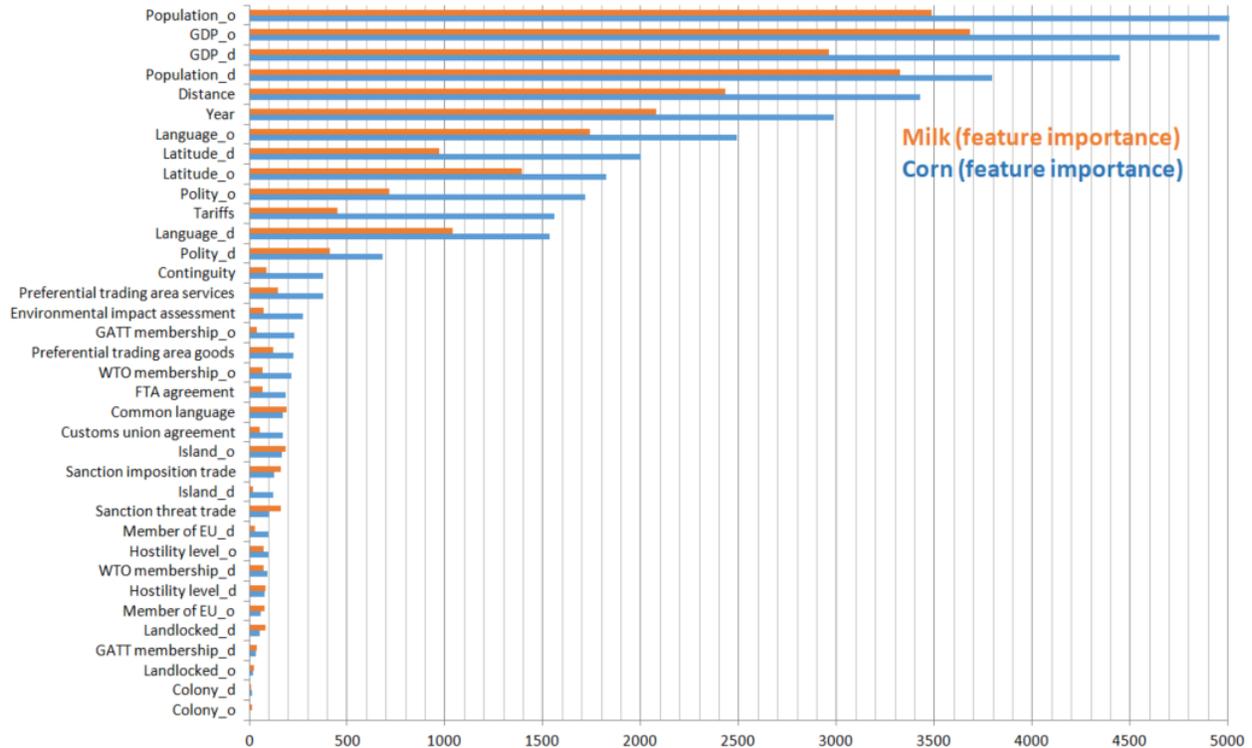

Figure 8: Economic variable importance for Milk and Corn (use case #1)

Our analysis of variables effecting milk's production can dictate policy decisions at the Dairy Margin Program (DMC) for instance, and associations of milk trade using our AR model. For example, through measures taken by USDA based on the Farm Bill (USDA, 2018), post Covid-19, dairy producers are granted a different set of agricultural loans based on the negative projections for production.

    Not all economic variables are influential during a pandemic, while other are more important to consider during conventional times (such as distance) and during policy making (such as

tariffs and production). In our boosting models, we considered the following economic variables (o refers to origin, and d refers to destination): *Distance, Population_d, GDP_d, GDP_o, Population_o, Colony_o, Colony_d, Landlocked_o, GATT membership_d, Landlocked_d, Member of EU_o, Hostility level_d, WTO membership_d, Hostility level_o, Member of EU_d, Sanction threat trade, Island_d, Sanction imposition trade, Island_o, Customs union agreement, Common language, FTA agreement, WTO membership_o, Preferential trading area goods, GATT membership_o, Environmental impact assessment, Preferential trading area services, Continguity, Polity_d, Language_d, Tariffs, Polity_o, Latitude_o, Latitude_d, Language_o, and Year* (all used in Figure 8).

By using AI methods, issues such as endogeneity are marginalized. Most traditional methods for high-dimensional statistics are based on the assumption that none of the regressors are correlated with the regression error (i.e. they are exogenous). Yet, endogeneity can arise incidentally from a large pool of regressors in cases of a high-dimensional regression. This leads to the inconsistency of the penalized least-squares method, and to possible false outcomes and sub-optimal policies. Our methods are not affected by this notion. However, in the next section, we present methods for validation of trade data (including anomaly detection), as well as other potential methods such as RL and causal learning that we aim to explore as part of future work.

## 5.2 Validation and Anomaly Detection

In the system developed, a validation and outlier detection module is deployed. The Validation engine presented in this section is based on work from Batarseh et al. (2017). The engine consists of two parts: flag validation, and anomaly (i.e. outlier) detection. We worked with USDA analysts to define the "acceptable" ranges for numbers within a sector, for example, if production of milk exceeds a certain amount, then our system should raise a "flag". For instance, for yearly values, a red flag means that the number is not within the accepted historical range, and it requires attention. Red flags are expected to be prominent during black swan events for example. An orange flag means the number is within the range but not the last 10 years, yellow means the value is within the last 10 years' range, but not 5 years range, then blue indicates 5 years range, and green indicates 3 years. The flag system is applied to livestock, in use case #2.

Another measure to identify outliers is using statistical methods: Median and Median Absolute Deviation Method (MAD). For this outlier detection method, the median of the residuals is calculated. Then, the difference is calculated between each historical value and this median. These differences are expressed as their absolute values, and a new median is calculated and multiplied by an empirically derived constant to yield the median absolute deviation (MAD). If a value is a certain number of MAD away from the median of the residuals, that value is classified as an outlier. The default threshold is 3 MAD. This method is generally more effective than the mean and standard deviation method for detecting outliers, but it can be too aggressive in classifying values that are not really extremely different. Also, if more than 50% of the data points have the same value, MAD is computed to be 0, so any value different from the residual median is classified as an outlier. MAD is calculated for univariate and multivariate outliers, using these two formulas; given a two dimensional paired set of data $(X_1, Y_1)$, $(X_2, Y_2)$ … $(X_n, Y_n)$:

$$MAD = median \left( |Xi - \tilde{X}| \right) \quad (6a)$$

$$Geometric \ (Multivariate) \ MAD = \left( median \left( |Xi - \tilde{X}| \right)^2 + median \left( |Yi - \tilde{Y}| \right)^2 \right)^{1/2} \quad (6b)$$

Use case #2 (the box below) is an example analysis of USDA's Economic Research Service (ERS) livestock data using validation and outliers (all code and data are available in a public Github repository – refer to the data availability section):

Usecase#2: U.S. livestock trade are collected from the National Agricultural Statistics Service (NASS) and the United States Department of Commerce (Foreign Trade Division). The flag and the anomaly detection modules were executed. Table 5 illustrates example results from the flag system. Table 6 presents example outlier values per commodity for the same dataset. All results are presented in Appendix 3. Such results are presented to a federal analyst for decision making and policy analysis.

Table 5: Flag Colors for Example USDA Data Series

| Data series | Unit | Value | Color |
|---|---|---|---|
| Pork, No Util Practice, No Prod Practice, No Physical Attribute, Exports | Million LBS, carcass-weight equivalent | 7148 | Red |
| Pork, No Util Practice, No Prod Practice, No Physical Attribute, Farm production | Million LBS, carcass-weight equivalent | 14.2 | Green |
| Pork, No Util Practice, No Prod Practice, No Physical Attribute, Imports | Million LBS, carcass-weight equivalent | 965 | Green |
| Pork, No Util Practice, No Prod Practice, No Physical Attribute, Per capita disappearance, Carcass weight | (LBS) Pounds | 64.1038 | Yellow |
| Pork, No Util Practice, No Prod Practice, No Physical Attribute, Per capita disappearance, Retail weight | (LBS) Pounds | 49.7445 | Yellow |

Table 6: Outlier Values (USDA data for Cattle and Hogs)

| Variable Description | Outlier Value | Time Value |
|---|---|---|
| U.S. Cattle imports, total | 250488 | 12/1/2014 |
| U.S. Cattle imports, 200 kilograms to less than 320 kilograms (705 pounds) | 134978 | 12/1/2014 |
| U.S. Cattle imports, 200 kilograms to less than 320 kilograms (705 pounds) | 142888 | 11/1/2014 |
| U.S. Cattle imports, total | 238125 | 12/1/2013 |
| U.S. Hog imports | 1105938 | 1/1/2008 |
| U.S. Cattle imports, total | 279413 | 11/1/2007 |
| U.S. Cattle imports, 320 kilograms or more, total | 178283 | 10/1/2007 |
| U.S. Cattle imports, total | 330750 | 11/1/2002 |
| U.S. Cattle exports, total | 122307 | 10/1/2001 |
| U.S. Cattle exports, other | 118790 | 10/1/2001 |
| U.S. Cattle exports, total | 129588 | 10/1/2000 |
| U.S. Cattle imports, 320 kilograms or more, for immediate slaughter | 138082 | 5/1/1996 |
| U.S. Cattle imports, 90 kilograms to less than 200 kilograms (440 pounds) | 196751 | 3/1/1995 |
| U.S. Cattle imports, total | 287087 | 3/1/1995 |
| U.S. Cattle imports, less than 320 kilograms, total | 252431 | 3/1/1995 |
| U.S. Cattle imports, total | 376650 | 3/1/1995 |

5.3 Future Work

As part of our future work, we aim to explore more AI methods for policy making. AI provides methods for causality and analysis of different scenarios and alternatives. For example, AI contextual reasoning could be deployed to define the *context* of an analysis (Stensrud et al., 2003). However, in most cases, it is highly challenging to define what *context* is. Context could be infinite (Bazire et al., 2005), and so data that ought to be collected to define a complete and universal context are also potentially infinite (Batarseh and Kulkarni, 2019).

Hanson et al. (2015) present an improvement on Ricardian trade models, in an approach that runs counter to recent theoretical research in trade; they rely on what they call "deep variables" (an equivalent to context variables in AI) to examine dynamics of comparative advantage. Context however, doesn't lead to understanding causality (a notion that is major in economics). An alternative to such modeling for big data sets are methods such as: RL, Causal Learning (CL), and other data-driven algorithms that can more explicitly point to causation.

For instance, the RL process is as follows: "the process starts from a start state and a number of actions are available to transform the start state to the next states. Similarly, for each of the next states in the search space, there are actions to transform them into yet other states. These transitions are given or are learned from carrying out actions in the environment. Rewards and punishments are given accordingly to the states depending on whether they are desired or undesired states. One version of RL, called Q-learning updates the state-action values, called Q-values, and causalities between actions and rewards are learned" (Ho, 2017). The RL process will find an optimal solution path to that state (Watkins, 1989). Q-learning is presented in the following formula:

$$V(s) \ = \ \text{Max of } a \ (R\,(s, a) \ + \ \gamma \, V\,(s')) \ \ (7)$$

Where, s = a particular state; a = action; s′ = state to which the model/agent goes from s, $\boldsymbol{\gamma}$ = discount factor; R (s, a) = a reward function which takes a state s and action a, and outputs a reward value; V (s) = value of being in a particular state.

Causal learning however relates more to cause and effect: Knowing what events cause what other events provides the information for an intelligent system to predict what may happen next so that it can better prepare and plan for the future (through deductive or inductive analysis). Causal learning also provides the means for it to solve problems by reasoning backward (i.e. counterfactuals): *If I desire trade outcome X to increase or decrease, what actions do I need to take?* (Ho, 2017). Such methods allow for acute and more detailed simulation of trade outcomes in alternative policy scenarios.

Additionally, in our future work, we aim to deploy HS4 and HS6 AR analysis to influence production decisions at farm level, provide commodity-based tariff insights, and other potential pointers to economic policies. A digital platform with "bigger" data shall provide acute associations on commodities' trade. HS4 or HS6 ARs for example, can point to some interesting pairing of countries and commodities. HS4 and HS6 codes will produce many more AR rules, which require more computational power, something we anticipate to do in the future.

## 6 Policy Implications

Methods presented in this paper provide ways of evaluating and managing policy scenarios. For instance, as presented in use cases 1 and 2, production of certain commodities can be hindered or amplified due to shifts in markets, changes in consumption, or other issues such as shipping and prices. Classification splits and gains lead to understanding these variables effects. Additionally, for instance: the amount of food available is calculated as the difference between available commodity supplies and nonfood use (i.e. disappearance) (USDA ERS, 2020):

*Total annual food supply of a commodity = supply (production + imports + beginning stocks) - disappearance (farm inputs + exports + ending stocks) (8)*

This (supply and use) allows analysts to produce reports that inform policy makers on commodities outlook, and so on policy decisions that influence variables per commodity. As presented in the two use cases and by using the presented EML and AR models, decision makers have better understanding of these variables, as they can predict them better, validate them and find outliers. Examples on policies across the world during outlier events include:

1. USA:

    a. The Covid-19 Food Assistance Program provides direct relief to producers who faced price declines and additional marketing costs due to the virus.

    b. The Office of Homeland Security has commenced a series of events to educate the Intelligence Community on threats to agriculture and the collection of information on emerging threats.

    c. Local and Regional Food Aid Procurement: On March 26, 2019, Foreign Agricultural Service (FAS) announced up to $15 million funding opportunity for Local and Regional Food Aid Procurement.

2. Canada:

    a. Farm Credit banks will receive support from the federal government that will allow for an additional $5.2 billion in lending capacity to producers, agribusinesses, and food processors (IMF, 2020).

3. South Africa and India:

    a. New funds are available to assist producers under stress, mainly small-scale farmers operating in the poultry, livestock, and vegetables sectors (IMF, 2020).

Policy on data sharing is critical to such studies. Encouraging government agencies to produce and share datasets on public repositories is essential for additional work in this context.

## 7   Conclusions

In recent years, there has been a challenge to achieve consensus at international bodies such as the United Nations, World Trade Organization, and others. For instance, the Doha round of multilateral trade negotiations initiated in 2001 has not been concluded as of today. Some have argued that G20, a group of 20 nations that account for much of world's GDP and population, has become increasingly assertive on multilateral policies. For instance, In Annex II of G20's Strategy for Global Trade Growth (SGTG, 2020), the G20 declares developing a world trade outlook *indicator*, and promoting further e-commerce development. That includes an initiative on an electronic World Trade Platform (e-WTP). The EML model illustrated in this paper can allow for contrasting between decisions made by G20 (i.e. mostly countries of cluster 1 in our analysis) versus engaging all countries like at the United Nations. The experimental work in this article indicates the high relevance of EML for predicting timely trade patterns with a greater accuracy than traditional approaches.

In our work, we show that transactional agricultural trade data could be used to enable insights into policy making such as: if the U.S. is exporting beef to Mexico $\Rightarrow$ Mexico will import less or no hay for animal feed (many other associations are presented). Similar to Market Basket Analysis (MBA) at grocery stores, trade trends for specific country-commodity pairs are directly affected by other "hidden" HS-2 commodity flows (confidence > 90%). Top world traders' imports and exports data (such as G7 or G20 countries) perform better training and scoring of AI models when compared to all international trade data ($R^2 > 69\%$). Countries of the world are best clustered in 6 groups based on their trade patterns. The 6-clusters model has low inter-cluster distances but high intra-cluster distances; and it influences boosting models that result in better predictions than traditional governmental outlooks. EML and AR models can have a direct effect on trade policy making during conventional times and black swan events in the U.S. and the world; our work evaluates and validates such scenarios.

Moreover, we show methods that analyze trade associations. AR is applied to HS2 commodity codes (i.e. oil seeds trade lead to cotton trade). EML models are based on clustering of countries and re-training of ML models based on that. Additionally, we present two use cases (using milk, corn, and livestock) for validation and outlier detection of economic commodity-specific variables, with application to agricultural economics. EML and AR models allow for simulation of trade outcomes in alternative policy scenarios. We offer a superior alternative to current approaches in public sector forecasting of trade flows. We rely on big data and AI – instead of complex behavioral models with assumptions solved by accessing information from a myriad of studies. We can AI methods on data to allow for alternative and robust specifications of complex economic relationships and trade policies.

## Authors Contributions:

**FB** defined the models, designed the experiments, developed the idea of the paper, presented the discussions and conclusions, and wrote the paper.

**GM** provided economic insights, designed the policy story, aided with data collection, and critically reviewed the paper.

**AM** developed the databases, wrote the scripts and code snippets, extracted results, and helped with technical writing.

**ZG** helped with writing literature review, helped with paper's idea design, and reviewed the paper.

## Data and Code Availability:

All data, code, scripts, and dashboards used in this study are available in this public repository:

https://github.com/ferasbatarseh/TradeAI

## Competing Interests:

No competing interests.

---

**Supplementary Material:**

**Appendix 1**

| Rule # | Antecedent | Consequent | Sum of Confidence |
|--------|------------|------------|-------------------|
| 1 | Cereals | Aircraft, spacecraft, and parts thereof | 946 |
| 2 | Cereals | Albuminoidal substances; modified starches; glues; enzymes | 946 |
| 3 | Cereals | Aluminum and articles thereof | 946 |
| 4 | Cereals | Animal or vegetable fats and oils and | 946 |

| | | their cleavage products; prepared edible fats; animal or vegetable waxes | |
|---|---|---|---|
| 5 | Cereals | Arms and ammunition; parts and accessories thereof | 946 |
| 6 | Cereals | Articles of apparel and clothing accessories, knitted or crocheted | 946 |
| 7 | Cereals | Articles of apparel and clothing accessories, not knitted or crocheted | 946 |
| 8 | Cereals | Articles of iron or steel | 946 |
| 9 | Cereals | Articles of leather; saddlery and harness; travel goods, handbags and similar containers; articles of animal gut (other than silkworm gut) | 946 |
| 10 | Cereals | Articles of stone, plaster, cement, asbestos, mica or similar materials | 946 |
| 11 | Products of the milling industry; malt; starches; inulin; wheat gluten | Fertilisers | 935 |
| 12 | Products of the milling industry; malt; starches; inulin; wheat gluten | Footwear, gaiters and the like; parts of such articles | 935 |
| 13 | Products of the milling industry; malt; starches; inulin; wheat gluten | Furniture; bedding, mattresses, mattress supports, cushions and similar stuffed furnishings; lamps and lighting fittings, not elsewhere specified or included; illuminated signs, illuminated nameplates | 935 |
| 14 | Products of the milling industry; malt; starches; inulin; wheat gluten | Furskins and artificial fur; manufactures thereof | 935 |
| 15 | Products of the milling industry; malt; starches; inulin; wheat gluten | Glass and glassware | 935 |
| 16 | Products of the milling industry; malt; starches; inulin; wheat gluten | Headgear and parts thereof | 935 |
| 17 | Products of the milling industry; malt; starches; inulin; wheat gluten | Impregnated, coated, covered or laminated textile fabrics; textile articles of a kind suitable for industrial use | 935 |
| 18 | Products of the milling | Inorganic chemicals; organic or inorganic | 935 |

| | industry; malt; starches; inulin; wheat gluten | compounds of precious metals, of rare earth metals, of radioactive elements or of isotopes | |
|---|---|---|---|
| 19 | Products of the milling industry; malt; starches; inulin; wheat gluten | Iron and steel | 935 |
| 20 | Products of the milling industry; malt; starches; inulin; wheat gluten | Knitted or crocheted fabrics | 935 |
| 21 | Animal or vegetable fats and oils and their cleavage products; prepared edible fats; animal or vegetable waxes | Cereals | 902 |
| 22 | Animal or vegetable fats and oils and their cleavage products; prepared edible fats; animal or vegetable waxes | Clocks and watches and parts thereof | 891 |
| 23 | Animal or vegetable fats and oils and their cleavage products; prepared edible fats; animal or vegetable waxes | Cocoa and cocoa preparations | 883.71 |
| 24 | Animal or vegetable fats and oils and their cleavage products; prepared edible fats; animal or vegetable waxes | Copper and articles thereof | 891 |
| 25 | Animal or vegetable fats and oils and their cleavage products; prepared edible fats; animal or vegetable waxes | Cork and articles of cork | 879.71 |
| 26 | Animal or vegetable fats and oils and their cleavage products; prepared edible fats; animal or vegetable waxes | Cotton | 891 |
| 27 | Animal or vegetable fats | Electrical machinery and equipment and | 891 |

| | | parts thereof; sound recorders and reproducers, television image and sound recorders and reproducers, and parts and accessories of such articles | |
|---|---|---|---|
| 28 | Animal or vegetable fats and oils and their cleavage products; prepared edible fats; animal or vegetable waxes | Essential oils and resinoids; perfumery, cosmetic or toilet preparations | 891 |
| 29 | Animal or vegetable fats and oils and their cleavage products; prepared edible fats; animal or vegetable waxes | Explosives; pyrotechnic products; matches; pyrophoric alloys; certain combustible preparations | 891 |
| 30 | Animal or vegetable fats and oils and their cleavage products; prepared edible fats; animal or vegetable waxes | Fertilizers | 891 |
| 31 | Oil seeds and oleaginous fruits; miscellaneous grains, seeds and fruit; industrial or medicinal plants; straw and fodder | Cereals | 935 |
| 32 | Oil seeds and oleaginous fruits; miscellaneous grains, seeds and fruit; industrial or medicinal plants; straw and fodder | Clocks and watches and parts thereof | 924 |
| 33 | Oil seeds and oleaginous fruits; miscellaneous grains, seeds and fruit; industrial or medicinal plants; straw and fodder | Cocoa and cocoa preparations | 916.43 |
| 34 | Oil seeds and oleaginous fruits; miscellaneous grains, seeds and fruit; industrial or medicinal plants; straw and fodder | Copper and articles thereof | 924 |
| 35 | Oil seeds and oleaginous | Cork and articles of cork | 912.29 |

| | fruits; miscellaneous grains, seeds and fruit; industrial or medicinal plants; straw and fodder | | |
|---|---|---|---|
| 36 | Oil seeds and oleaginous fruits; miscellaneous grains, seeds and fruit; industrial or medicinal plants; straw and fodder | Cotton | 924 |
| 37 | Oil seeds and oleaginous fruits; miscellaneous grains, seeds and fruit; industrial or medicinal plants; straw and fodder | Electrical machinery and equipment and parts thereof; sound recorders and reproducers, television image and sound recorders and reproducers, and parts and accessories of such articles | 924 |
| 38 | Oil seeds and oleaginous fruits; miscellaneous grains, seeds and fruit; industrial or medicinal plants; straw and fodder | Essential oils and resinoids; perfumery, cosmetic or toilet preparations | 924 |
| 39 | Oil seeds and oleaginous fruits; miscellaneous grains, seeds and fruit; industrial or medicinal plants; straw and fodder | Explosives; pyrotechnic products; matches; pyrophoric alloys; certain combustible preparations | 924 |
| 40 | Oil seeds and oleaginous fruits; miscellaneous grains, seeds and fruit; industrial or medicinal plants; straw and fodder | Fertilizers | 924 |
| 41 | Lac; gums, resins and other vegetable saps and extracts | Aircraft, spacecraft, and parts thereof | 913 |
| 42 | Lac; gums, resins and other vegetable saps and extracts | Albuminoidal substances; modified starches; glues; enzymes | 913 |
| 43 | Lac; gums, resins and other vegetable saps and extracts | Aluminum and articles thereof | 913 |
| 44 | Lac; gums, resins and other vegetable saps and extracts | Animal or vegetable fats and oils and their cleavage products; prepared edible fats; animal or vegetable waxes | 913 |
| 45 | Lac; gums, resins and other vegetable saps and extracts | Arms and ammunition; parts and accessories thereof | 913 |

| 46 | Lac; gums, resins and other vegetable saps and extracts | Articles of apparel and clothing accessories, knitted or crocheted | 913 |
|---|---|---|---|
| 47 | Lac; gums, resins and other vegetable saps and extracts | Articles of apparel and clothing accessories, not knitted or crocheted | 913 |
| 48 | Lac; gums, resins and other vegetable saps and extracts | Articles of iron or steel | 913 |
| 49 | Lac; gums, resins and other vegetable saps and extracts | Articles of leather; saddlery and harness; travel goods, handbags and similar containers; articles of animal gut (other than silkworm gut) | 913 |
| 50 | Lac; gums, resins and other vegetable saps and extracts | Articles of stone, plaster, cement, asbestos, mica or similar materials | 913 |
| 51 | Vegetable plaiting materials; vegetable products not elsewhere specified or included | Aircraft, spacecraft, and parts thereof | 902 |
| 52 | Vegetable plaiting materials; vegetable products not elsewhere specified or included | Albuminoidal substances; modified starches; glues; enzymes | 902 |
| 53 | Vegetable plaiting materials; vegetable products not elsewhere specified or included | Aluminum and articles thereof | 902 |
| 54 | Vegetable plaiting materials; vegetable products not elsewhere specified or included | Animal or vegetable fats and oils and their cleavage products; prepared edible fats; animal or vegetable waxes | 902 |
| 55 | Vegetable plaiting materials; vegetable products not elsewhere specified or included | Arms and ammunition; parts and accessories thereof | 902 |
| 56 | Vegetable plaiting materials; vegetable products not elsewhere specified or included | Articles of apparel and clothing accessories, knitted or crocheted | 902 |
| 57 | Vegetable plaiting materials; vegetable products not elsewhere specified or included | Articles of apparel and clothing accessories, not knitted or crocheted | 902 |

| 58 | Vegetable plaiting materials; vegetable products not elsewhere specified or included | Articles of iron or steel | 902 |
|---|---|---|---|
| 59 | Vegetable plaiting materials; vegetable products not elsewhere specified or included | Articles of leather; saddlery and harness; travel goods, handbags and similar containers; articles of animal gut (other than silkworm gut) | 902 |
| 60 | Vegetable plaiting materials; vegetable products not elsewhere specified or included | Articles of stone, plaster, cement, asbestos, mica or similar materials | 902 |
| 61 | Sugars and sugar confectionery | Cocoa and cocoa preparations | 861.89 |
| 62 | Sugars and sugar confectionery | Copper and articles thereof | 869 |
| 63 | Sugars and sugar confectionery | Cork and articles of cork | 857.99 |
| 64 | Sugars and sugar confectionery | Cotton | 869 |
| 65 | Sugars and sugar confectionery | Electrical machinery and equipment and parts thereof; sound recorders and reproducers, television image and sound recorders and reproducers, and parts and accessories of such articles | 869 |
| 66 | Sugars and sugar confectionery | Essential oils and resinoids; perfumery, cosmetic or toilet preparations | 869 |
| 67 | Sugars and sugar confectionery | Explosives; pyrotechnic products; matches; pyrophoric alloys; certain combustible preparations | 869 |
| 68 | Sugars and sugar confectionery | Fertilizers | 869 |
| 69 | Sugars and sugar confectionery | Footwear, gaiters and the like; parts of such articles | 869 |
| 70 | Sugars and sugar confectionery | Furniture; bedding, mattresses, mattress supports, cushions and similar stuffed furnishings; lamps and lighting fittings, not elsewhere specified or included; illuminated signs, illuminated nameplates | 869 |

| 71 | Preparations of vegetables, fruit, nuts or other parts of plants | Silk | 836 |
|---|---|---|---|
| 72 | Preparations of vegetables, fruit, nuts or other parts of plants | Soap, organic surface-active agents, washing preparations, lubricating preparations, artificial waxes, prepared waxes, polishing or scouring preparations, candles and similar articles, modeling pastes | 836 |
| 73 | Preparations of vegetables, fruit, nuts or other parts of plants | Special woven fabrics; tufted textile fabrics; lace; tapestries; trimmings; embroidery | 836 |
| 74 | Preparations of vegetables, fruit, nuts or other parts of plants | Sugars and sugar confectionery | 847 |
| 75 | Preparations of vegetables, fruit, nuts or other parts of plants | Tanning or dyeing extracts; tannins and their derivatives; dyes, pigments and other colouring matter; paints and varnishes; putty and other mastics; inks | 836 |
| 76 | Preparations of vegetables, fruit, nuts or other parts of plants | Tin and articles thereof | 836 |
| 77 | Preparations of vegetables, fruit, nuts or other parts of plants | Tobacco and manufactured tobacco substitutes | 760 |
| 78 | Preparations of vegetables, fruit, nuts or other parts of plants | Tools, implements, cutlery, spoons and forks, of base metal; parts thereof of base metal | 836 |
| 79 | Preparations of vegetables, fruit, nuts or other parts of plants | Toys, games and sports requisites; parts and accessories thereof | 836 |
| 80 | Preparations of vegetables, fruit, nuts or other parts of plants | Umbrellas, sun umbrellas, walking sticks, seat sticks, whips, riding crops and parts thereof | 836 |
| 81 | Cocoa and cocoa preparations | Fur skins and artificial fur; manufactures thereof | 858 |
| 82 | Cocoa and cocoa preparations | Glass and glassware | 858 |
| 83 | Cocoa and cocoa | Headgear and parts thereof | 858 |
| | preparations | | |
| 84 | Cocoa and cocoa preparations | Impregnated, coated, covered or laminated textile fabrics; textile articles of a kind suitable for industrial use | 858 |
| 85 | Cocoa and cocoa preparations | Inorganic chemicals; organic or inorganic compounds of precious metals, of rare earth metals, of radioactive elements or of isotopes | 858 |
| 86 | Cocoa and cocoa preparations | Iron and steel | 858 |
| 87 | Cocoa and cocoa preparations | Knitted or crocheted fabrics | 858 |
| 88 | Cocoa and cocoa preparations | Lac; gums, resins and other vegetable saps and extracts | 869 |
| 89 | Cocoa and cocoa preparations | Lead and articles thereof | 780 |
| 90 | Cocoa and cocoa preparations | Manmade filaments | 858 |
| 91 | Preparations of cereals, flour, starch or milk; pastrycooks' products | Musical instruments; parts and accessories of such articles | 847 |
| 92 | Preparations of cereals, flour, starch or milk; pastrycooks' products | Natural or cultured pearls, precious or semiprecious stones, precious metals, metals clad with precious metal, and articles thereof; imitation jewelry; coin | 847 |
| 93 | Preparations of cereals, flour, starch or milk; pastrycooks' products | Nickel and articles thereof | 847 |
| 94 | Preparations of cereals, flour, starch or milk; pastrycooks' products | Nuclear reactors, boilers, machinery and mechanical appliances; parts thereof | 847 |
| 95 | Preparations of cereals, flour, starch or milk; pastrycooks' products | Oil seeds and oleaginous fruits; miscellaneous grains, seeds and fruit; industrial or medicinal plants; straw and fodder | 858 |
| 96 | Preparations of cereals, flour, starch or milk; pastrycooks' products | Optical, photographic, cinematographic, measuring, checking, precision, medical or surgical instruments and apparatus; parts and accessories thereof | 847 |

| 97 | Preparations of cereals, flour, starch or milk; pastrycooks' products | Ores, slag and ash | 840.03 |
| 98 | Preparations of cereals, flour, starch or milk; pastrycooks' products | Organic chemicals | 847 |
| 99 | Beverages, spirits and vinegar | Explosives; pyrotechnic products; matches; pyrophoric alloys; certain combustible preparations | 814 |
| 100 | Beverages, spirits and vinegar | Fertilizers | 814 |

---

## Appendix 2

| Country | k=2 | k=3 | k=4 | k=5 | k=6 | k=7 | k=8 | k=9 | k=10 |
|---|---|---|---|---|---|---|---|---|---|
| Argentina | 1 | 2 | 2 | 2 | 2 | 5 | 1 | 3 | 1 |
| Armenia | 0 | 1 | 0 | 4 | 3 | 2 | 6 | 8 | 9 |
| Australia | 1 | 2 | 1 | 1 | 5 | 4 | 4 | 1 | 3 |
| Bahrain | 0 | 0 | 0 | 3 | 0 | 2 | 6 | 0 | 0 |
| Belize | 0 | 1 | 3 | 0 | 4 | 1 | 3 | 5 | 2 |
| Benin | 0 | 1 | 3 | 4 | 3 | 1 | 7 | 4 | 4 |
| Bolivia | 0 | 0 | 0 | 3 | 0 | 2 | 6 | 0 | 0 |
| Botswana | 0 | 0 | 0 | 3 | 0 | 2 | 6 | 0 | 0 |
| Brazil | 1 | 2 | 1 | 1 | 5 | 4 | 4 | 1 | 3 |
| Brunei Darussalam | 0 | 1 | 0 | 4 | 3 | 1 | 6 | 2 | 9 |
| Burkina Faso | 0 | 1 | 3 | 4 | 3 | 1 | 7 | 4 | 4 |
| Burundi | 0 | 1 | 3 | 0 | 4 | 6 | 3 | 5 | 2 |
| Cabo Verde | 0 | 1 | 3 | 0 | 4 | 6 | 3 | 5 | 2 |
| Cameroon | 0 | 1 | 0 | 4 | 3 | 1 | 0 | 2 | 9 |
| Canada | 1 | 2 | 1 | 1 | 5 | 4 | 4 | 1 | 3 |
| Chad | 0 | 1 | 3 | 0 | 4 | 1 | 7 | 4 | 2 |
| Chile | 1 | 2 | 2 | 2 | 2 | 5 | 1 | 3 | 1 |
| China | 1 | 2 | 1 | 1 | 1 | 3 | 2 | 7 | 7 |
| Chinese Taipei | 1 | 2 | 1 | 1 | 5 | 4 | 4 | 1 | 3 |
| Colombia | 1 | 2 | 2 | 2 | 2 | 5 | 1 | 3 | 1 |
| Costa Rica | 0 | 0 | 2 | 3 | 0 | 0 | 5 | 6 | 8 |

| Country | | | | | | | | | |
|---|---|---|---|---|---|---|---|---|---|
| Cuba | 0 | 0 | 0 | 3 | 0 | 0 | 5 | 6 | 8 |
| Dominican Republic | 0 | 0 | 2 | 3 | 0 | 0 | 5 | 6 | 8 |
| Ecuador | 0 | 0 | 2 | 3 | 0 | 0 | 5 | 6 | 8 |
| Egypt | 1 | 2 | 2 | 2 | 2 | 5 | 1 | 3 | 1 |
| El Salvador | 0 | 0 | 0 | 3 | 0 | 0 | 5 | 6 | 8 |
| European Union | 1 | 2 | 1 | 1 | 1 | 3 | 2 | 7 | 7 |
| Georgia | 0 | 0 | 0 | 3 | 0 | 2 | 6 | 8 | 0 |
| Haiti | 0 | 1 | 0 | 4 | 3 | 2 | 6 | 2 | 9 |
| Hong Kong, China | 1 | 2 | 1 | 1 | 5 | 4 | 4 | 1 | 3 |
| India | 1 | 2 | 1 | 1 | 5 | 4 | 4 | 1 | 3 |
| Indonesia | 1 | 2 | 1 | 1 | 5 | 4 | 4 | 1 | 3 |
| Israel | 1 | 2 | 2 | 2 | 2 | 5 | 1 | 3 | 1 |
| Jamaica | 0 | 0 | 0 | 3 | 0 | 2 | 6 | 0 | 0 |
| Japan | 1 | 2 | 1 | 1 | 1 | 3 | 2 | 7 | 7 |
| Jordan | 0 | 0 | 2 | 3 | 0 | 0 | 5 | 6 | 8 |
| Kazakhstan | 1 | 0 | 2 | 2 | 2 | 5 | 1 | 3 | 6 |
| S. Korea | 1 | 2 | 1 | 1 | 5 | 4 | 4 | 1 | 3 |
| Kuwait | 1 | 0 | 2 | 2 | 2 | 0 | 1 | 3 | 6 |
| Kyrgyz Republic | 0 | 0 | 0 | 4 | 0 | 2 | 6 | 8 | 0 |
| Lao | 0 | 0 | 0 | 3 | 0 | 2 | 5 | 0 | 0 |
| Macao, China | 0 | 0 | 0 | 3 | 0 | 0 | 6 | 0 | 0 |
| Madagascar | 0 | 1 | 0 | 4 | 3 | 1 | 7 | 2 | 9 |
| Malawi | 0 | 1 | 3 | 4 | 3 | 1 | 7 | 4 | 4 |
| Malaysia | 1 | 2 | 1 | 1 | 5 | 4 | 4 | 1 | 3 |
| Maldives | 0 | 1 | 3 | 0 | 4 | 1 | 7 | 4 | 2 |
| Mali | 0 | 1 | 3 | 4 | 3 | 1 | 7 | 4 | 4 |
| Mauritania | 0 | 1 | 3 | 4 | 3 | 1 | 7 | 4 | 4 |
| Mauritius | 0 | 0 | 0 | 4 | 0 | 2 | 6 | 8 | 0 |
| Mexico | 1 | 2 | 1 | 1 | 5 | 4 | 4 | 1 | 3 |
| Moldova | 0 | 0 | 0 | 3 | 0 | 2 | 6 | 8 | 0 |
| Mongolia | 0 | 1 | 0 | 4 | 3 | 1 | 6 | 8 | 9 |
| Montenegro | 0 | 1 | 3 | 0 | 4 | 1 | 7 | 8 | 2 |
| Morocco | 1 | 0 | 2 | 2 | 2 | 5 | 1 | 3 | 1 |
| Mozambique | 0 | 0 | 0 | 4 | 0 | 2 | 0 | 2 | 9 |
| Myanmar | 0 | 0 | 2 | 3 | 2 | 0 | 5 | 6 | 8 |
| Namibia | 0 | 0 | 0 | 3 | 0 | 2 | 6 | 0 | 0 |
| Nepal | 0 | 0 | 0 | 3 | 0 | 2 | 5 | 6 | 0 |
| New Zealand | 1 | 0 | 2 | 2 | 2 | 5 | 1 | 3 | 6 |

| | | | | | | | | |
|---|---|---|---|---|---|---|---|---|
| **Nicaragua** | 0 | 0 | 0 | 4 | 0 | 2 | 6 | 2 | 9 |
| **Niger** | 0 | 1 | 3 | 0 | 3 | 1 | 7 | 4 | 4 |
| **Nigeria** | 1 | 0 | 2 | 2 | 2 | 5 | 1 | 3 | 1 |
| **North Macedonia** | 0 | 0 | 0 | 3 | 0 | 2 | 6 | 8 | 0 |
| **Norway** | 1 | 2 | 2 | 2 | 2 | 5 | 1 | 3 | 1 |
| **Pakistan** | 1 | 0 | 2 | 2 | 2 | 5 | 1 | 3 | 8 |
| **Panama** | 0 | 0 | 0 | 3 | 0 | 0 | 5 | 0 | 0 |
| **Paraguay** | 0 | 0 | 0 | 3 | 0 | 0 | 5 | 6 | 0 |
| **Peru** | 1 | 0 | 2 | 2 | 2 | 5 | 1 | 3 | 1 |
| **Philippines** | 1 | 2 | 2 | 2 | 2 | 5 | 4 | 1 | 1 |
| **Qatar** | 1 | 0 | 2 | 2 | 2 | 5 | 1 | 3 | 6 |
| **Rwanda** | 0 | 1 | 3 | 4 | 3 | 1 | 7 | 4 | 4 |
| **Saudi Arabia** | 1 | 2 | 1 | 1 | 5 | 4 | 4 | 1 | 3 |
| **Senegal** | 0 | 1 | 0 | 4 | 3 | 2 | 0 | 2 | 9 |
| **Seychelles** | 0 | 1 | 3 | 0 | 4 | 6 | 3 | 5 | 5 |
| **Sierra Leone** | 0 | 1 | 0 | 4 | 3 | 1 | 6 | 8 | 9 |
| **Singapore** | 1 | 2 | 1 | 1 | 5 | 4 | 4 | 1 | 3 |
| **South Africa** | 1 | 2 | 2 | 2 | 2 | 5 | 1 | 3 | 1 |
| **Sri Lanka** | 0 | 0 | 2 | 3 | 0 | 0 | 5 | 6 | 8 |
| **Suriname** | 0 | 1 | 3 | 0 | 4 | 1 | 7 | 4 | 2 |
| **Switzerland** | 1 | 2 | 1 | 1 | 5 | 4 | 4 | 1 | 3 |
| **Tanzania** | 0 | 0 | 0 | 4 | 3 | 2 | 0 | 2 | 9 |
| **Thailand** | 1 | 2 | 1 | 1 | 5 | 4 | 4 | 1 | 3 |
| **Togo** | 0 | 1 | 3 | 4 | 3 | 1 | 7 | 4 | 4 |
| **Tonga** | 0 | 1 | 3 | 0 | 4 | 6 | 3 | 5 | 5 |
| **Tunisia** | 0 | 0 | 2 | 3 | 0 | 0 | 5 | 6 | 8 |
| **Uganda** | 0 | 1 | 0 | 4 | 3 | 1 | 0 | 2 | 9 |
| **Ukraine** | 1 | 0 | 2 | 2 | 2 | 5 | 1 | 3 | 1 |
| **United States of America** | 1 | 2 | 1 | 1 | 1 | 3 | 2 | 7 | 7 |
| **Uruguay** | 0 | 0 | 0 | 3 | 0 | 0 | 5 | 6 | 0 |
| **Venezuela** | 1 | 0 | 2 | 2 | 2 | 5 | 1 | 3 | 1 |
| **Zambia** | 0 | 0 | 0 | 4 | 0 | 2 | 0 | 2 | 9 |
| **Zimbabwe** | 0 | 0 | 0 | 4 | 0 | 2 | 0 | 2 | 9 |

**Appendix 3**

| Commodity Description | Statistical Type | Unit | Value | Flag Color |
|---|---|---|---|---|
| Chickens, Broilers, All utilization practices, All production practices, No Physical Attribute, Feed price ratio | Feed price ratio | Lb grower feed to lb live weight | 4.5 | Yellow |
| Hogs, All utilization practices, All production practices, No Physical Attribute, Feed price ratio | Feed price ratio | Bu corn to cwt live weight | 12.3 | Yellow |
| Milk, Class IV milk, All production practices, No Physical Attribute, Price received | Price received | $ / cwt | 16.8 | Yellow |
| Milk, All utilization practices, All production practices, No Physical Attribute, Price received | Price received | $ / cwt | 18.8 | Yellow |
| Pork, No Util Practice, No Prod Practice, No Physical Attribute, Per capita disappearance, retail weight | Per capita disappearance, retail weight | (LBS) Pounds | 49.7445 | Yellow |
| Pork, No Util Practice, No Prod Practice, No Physical Attribute, Per capita disappearance, carcass weight | Per capita disappearance, carcass weight | (LBS) Pounds | 64.1038 | Yellow |
| Beef, No Util Practice, No Prod Practice, No Physical Attribute, Imports | Imports | Million LBS, carcass-weight equivalent | 2903.8 | Yellow |
| Beef, No Util Practice, No Prod Practice, No Physical Attribute, Imports | Imports | Million LBS, carcass-weight equivalent | 2960 | Yellow |
| Total red meat, No Util Practice, No Prod Practice, No Physical Attribute, Imports | Imports | Million LBS, carcass-weight equivalent | 3993 | Yellow |
| Total red meat, No Util Practice, No Prod Practice, No Physical Attribute, Imports | Imports | Million LBS, carcass-weight equivalent | 4196 | Yellow |
| Chickens, Broilers, No Util Practice, No Prod Practice, No Physical Attribute, Exports | Exports | Million LBS | 7250 | Yellow |
| Total poultry, No Util Practice, No Prod Practice, No Physical Attribute, Exports | Exports | Million LBS | 7952 | Yellow |
| Total poultry, No Util Practice, No Prod Practice, No Physical Attribute, Exports | Exports | Million LBS | 8022.9 | Yellow |
| Milk, All utilization practices, All production practices, No Physical Attribute, Domestic commercial use, fat basis | Domestic commercial use, fat basis | Billion lbs, milk equivalent | 216.9341 | Red |
| Milk, All utilization practices, All production practices, No Physical Attribute, Domestic commercial use, fat basis | Domestic commercial use, fat basis | Billion lbs, milk equivalent | 217.8379 | Red |
| Milk, All farm milk marketed, All production practices, No Physical Attribute, Marketings | Marketings | Billion lbs | 221.3392 | Red |
| Milk, All farm milk marketed, All production practices, No Physical Attribute, Marketings | Marketings | Billion lbs | 221.7479 | Red |
| Milk, All utilization practices, All production practices, No Physical Attribute, Total commercial supply, skim-solid basis | Total commercial supply, skim-solid basis | Billion lbs, milk equivalent | 236.9179 | Red |
| Milk, All utilization practices, All production practices, No Physical Attribute, Total commercial supply, skim-solid basis | Total commercial supply, skim-solid basis | Billion lbs, milk equivalent | 237.2613 | Red |
| Milk, All utilization practices, All production practices, No Physical Attribute, Total commercial supply, fat basis | Total commercial supply, fat basis | Billion lbs, milk equivalent | 241.3279 | Red |
| Milk, All utilization practices, All production practices, No Physical Attribute, Total commercial supply, fat basis | Total commercial supply, fat basis | Billion lbs, milk equivalent | 241.653 | Red |
| Pork, No Util Practice, No Prod Practice, No Physical Attribute, Exports | Exports | Million LBS, carcass-weight equivalent | 6675 | Red |
| Pork, No Util Practice, No Prod Practice, No Physical Attribute, Exports | Exports | Million LBS, carcass- | 7148 | Red |

| | | weight equivalent | | |
|---|---|---|---|---|
| Eggs and egg products, No Util Practice, No Prod Practice, No Physical Attribute, Total disappearance | Total disappearance | Million dozen, shell-egg equivalent | 8085 | Red |
| Eggs and egg products, No Util Practice, No Prod Practice, No Physical Attribute, Total disappearance | Total disappearance | Million dozen, shell-egg equivalent | 8116.522 | Red |
| Eggs and egg products, No Util Practice, No Prod Practice, No Physical Attribute, Table production | Table production | Million dozen, shell-egg equivalent | 8270 | Red |
| Eggs and egg products, No Util Practice, No Prod Practice, No Physical Attribute, Table production | Table production | Million dozen, shell-egg equivalent | 8312.33 | Red |
| Eggs and egg products, No Util Practice, No Prod Practice, No Physical Attribute, Total supply | Total supply | Million dozen, shell-egg equivalent | 9550 | Red |
| Eggs and egg products, No Util Practice, No Prod Practice, No Physical Attribute, Total supply | Total supply | Million dozen, shell-egg equivalent | 9636.422 | Red |
| Total red meat, No Util Practice, No Prod Practice, No Physical Attribute, Exports | Exports | Million LBS, carcass-weight equivalent | 9927 | Red |
| Total red meat, No Util Practice, No Prod Practice, No Physical Attribute, Exports | Exports | Million LBS, carcass-weight equivalent | 10050.1 | Red |
| Butter, Wholesale market, All production practices, No Physical Attribute, Wholesale price | Wholesale price | $ / lb | 1.4101 | Orange |
| Veal, No Util Practice, No Prod Practice, No Physical Attribute, Exports | Exports | Million LBS, carcass-weight | 0 | Green |

| | | equivalent | | |
|---|---|---|---|---|
| Veal, No Util Practice, No Prod Practice, No Physical Attribute, Imports | Imports | Million LBS, carcass-weight equivalent | 0 | Green |
| Turkeys, No Util Practice, No Prod Practice, No Physical Attribute, Condemnations | Condemnations | Million LBS | 0 | Green |
| Milk, All utilization practices, All production practices, No Physical Attribute, Net removals,  fat basis | Net removals, fat basis | Billion lbs, milk equivalent | 0 | Green |
| Milk, All utilization practices, All production practices, No Physical Attribute, Net removals,  skim-solid basis | Net removals, skim-solid basis | Billion lbs, milk equivalent | 0 | Green |
| Milk, All utilization practices, All production practices, No Physical Attribute, Net removals,  skim-solid basis | Net removals, skim-solid basis | Billion lbs, milk equivalent | 0.1196 | Green |
| Veal, No Util Practice, No Prod Practice, No Physical Attribute, Per capita disappearance,  retail weight | Per capita disappearance, retail weight | (LBS) Pounds | 0.2129 | Green |
| Veal, No Util Practice, No Prod Practice, No Physical Attribute, Per capita disappearance,  carcass weight | Per capita disappearance, carcass weight | (LBS) Pounds | 0.2565 | Green |
| Whey,  Dry, Wholesale market, All production practices, Condition, Wholesale price | Wholesale price | $ / lb | 0.3795 | Green |
| Whey,  Dry, Wholesale market, All production practices, Condition, Wholesale price | Wholesale price | $ / LB | 0.38 | Green |
| Milk,  Dry,  Nonfat, Wholesale market, All production practices, Condition, Wholesale price | Wholesale price | $ / LB | 0.9393 | Green |
| Milk, Fed to calves and home consumption, All production practices, No Physical Attribute, Farm use | Farm use | Billion lbs | 0.979 | Green |
| Lamb and mutton, No Util Practice, No Prod Practice, No Physical Attribute, Per | Per capita disappearance, | (LBS) Pounds | 1.1062 | Green |

| capita disappearance,  retail weight | retail weight | | | |
|---|---|---|---|---|
| Lamb and mutton, No Util Practice, No Prod Practice, No Physical Attribute, Per capita disappearance,  retail weight | Per capita disappearance, retail weight | (LBS) Pounds | 1.1177 | Green |
| Lamb and mutton, No Util Practice, No Prod Practice, No Physical Attribute, Per capita disappearance,  carcass weight | Per capita disappearance, carcass weight | (LBS) Pounds | 1.2429 | Green |
| Lamb and mutton, No Util Practice, No Prod Practice, No Physical Attribute, Per capita disappearance,  carcass weight | Per capita disappearance, carcass weight | (LBS) Pounds | 1.2559 | Green |
| Chickens,  Other, No Util Practice, No Prod Practice, No Physical Attribute, Per capita disappearance,  carcass weight | Per capita disappearance, carcass weight | (LBS) Pounds | 1.4276 | Green |
| Chickens,  Other, No Util Practice, No Prod Practice, No Physical Attribute, Per capita disappearance,  retail weight | Per capita disappearance, retail weight | (LBS) Pounds | 1.4276 | Green |
| Chickens,  Other, No Util Practice, No Prod Practice, No Physical Attribute, Per capita disappearance,  carcass weight | Per capita disappearance, carcass weight | (LBS) Pounds | 1.4394 | Green |
| Chickens,  Other, No Util Practice, No Prod Practice, No Physical Attribute, Per capita disappearance,  retail weight | Per capita disappearance, retail weight | (LBS) Pounds | 1.4394 | Green |
| Cheese,  Cheddar, Wholesale market, All production practices, No Physical Attribute, Wholesale price | Wholesale price | $ / LB | 1.71 | Green |
| Milk, All utilization practices, All production practices, No Physical Attribute, Feed price ratio | Feed price ratio | Lb 16% mixed dairy feed to lb whole milk | 2.37 | Green |
| Chickens,  Other, No Util Practice, No Prod Practice, No Physical Attribute, Imports | Imports | Million LBS | 2.668 | Green |
| Chickens,  Other, No Util Practice, No Prod Practice, No Physical Attribute, Imports | Imports | Million LBS | 3 | Green |
| Lamb and mutton, No Util Practice, No Prod Practice, No Physical Attribute, Farm production | Farm production | Million LBS, carcass-weight equivalent | 4.9 | Green |
| Veal, No Util Practice, No Prod Practice, No Physical Attribute, Farm production | Farm production | Million LBS, carcass-weight equivalent | 5 | Green |
| Chickens,  Other, No Util Practice, No Prod Practice, No Physical Attribute, Ending stocks | Ending stocks | Million LBS | 5 | Green |
| Veal, No Util Practice, No Prod Practice, No Physical Attribute, Farm production | Farm production | Million LBS, carcass-weight equivalent | 5.1 | Green |
| Lamb and mutton, No Util Practice, No Prod Practice, No Physical Attribute, Farm production | Farm production | Million LBS, carcass-weight equivalent | 5.2 | Green |
| Chickens,  Other, No Util Practice, No Prod Practice, No Physical Attribute, Beginning stocks | Beginning stocks | Million LBS | 5.433 | Green |
| Chickens,  Other, No Util Practice, No Prod Practice, No Physical Attribute, Beginning stocks | Beginning stocks | Million LBS | 5.615 | Green |
| Milk, All utilization practices, All production practices, No Physical Attribute, Imports,  skim-solid basis | Imports,  skim-solid basis | Billion lbs, milk equivalent | 5.7188 | Green |
| Milk, All utilization practices, All production practices, No Physical Attribute, Imports,  fat basis | Imports,  fat basis | Billion lbs, milk equivalent | 6.58 | Green |
| Milk, All utilization practices, All production practices, No Physical Attribute, Imports,  fat basis | Imports,  fat basis | Billion lbs, milk equivalent | 6.6756 | Green |
| Turkeys, All utilization practices, All production practices, No Physical Attribute, Feed price ratio | Feed price ratio | Lb grower feed to lb live weight | 6.9 | Green |
| Veal, No Util Practice, No Prod Practice, No Physical Attribute, Ending stocks | Ending stocks | Million LBS, carcass-weight equivalent | 7 | Green |
| Lamb and mutton, No Util Practice, No Prod Practice, No Physical Attribute, Exports | Exports | Million LBS, carcass-weight | 7 | Green |

| | | equivalent | | |
|---|---|---|---|---|
| Veal, No Util Practice, No Prod Practice, No Physical Attribute, Beginning stocks | Beginning stocks | Million LBS, carcass-weight equivalent | 9 | Green |
| Veal, No Util Practice, No Prod Practice, No Physical Attribute, Ending stocks | Ending stocks | Million LBS, carcass-weight equivalent | 9 | Green |
| Milk, All utilization practices, All production practices, No Physical Attribute, Commercial exports,  fat basis | Commercial exports,  fat basis | Billion lbs, milk equivalent | 9.1629 | Green |
| Milk, All utilization practices, All production practices, No Physical Attribute, Beginning commercial stocks, skim-solid basis | Beginning commercial stocks,  skim-solid basis | Billion lbs, milk equivalent | 10.2 | Green |
| Milk, All utilization practices, All production practices, No Physical Attribute, Beginning commercial stocks, skim-solid basis | Beginning commercial stocks,  skim-solid basis | Billion lbs, milk equivalent | 10.2033 | Green |
| Milk, All utilization practices, All production practices, No Physical Attribute, Ending commercial stocks, skim-solid basis | Ending commercial stocks,  skim-solid basis | Billion lbs, milk equivalent | 10.3 | Green |
| Milk, All utilization practices, All production practices, No Physical Attribute, Ending commercial stocks, skim-solid basis | Ending commercial stocks,  skim-solid basis | Billion lbs, milk equivalent | 11.3 | Green |
| Turkeys, No Util Practice, No Prod Practice, No Physical Attribute, Imports | Imports | Million LBS | 13 | Green |
| Milk, All utilization practices, All production practices, No Physical Attribute, Beginning commercial stocks, fat basis | Beginning commercial stocks,  fat basis | Billion lbs, milk equivalent | 13 | Green |
| Milk, All utilization practices, All production practices, No Physical Attribute, Beginning commercial stocks, fat basis | Beginning commercial stocks,  fat basis | Billion lbs, milk equivalent | 13.6381 | Green |
| Pork, No Util Practice, No Prod Practice, No Physical Attribute, Farm | Farm production | Million LBS, carcass- | 13.7 | Green |
| production | | weight equivalent | | |
| Pork, No Util Practice, No Prod Practice, No Physical Attribute, Farm production | Farm production | Million LBS, carcass-weight equivalent | 14.2 | Green |
| Turkeys, No Util Practice, No Prod Practice, No Physical Attribute, Imports | Imports | Million LBS | 16 | Green |
| Turkeys, No Util Practice, No Prod Practice, No Physical Attribute, Per capita disappearance,  carcass weight | Per capita disappearance, carcass weight | (LBS) Pounds | 16.0712 | Green |
| Turkeys, No Util Practice, No Prod Practice, No Physical Attribute, Per capita disappearance,  retail weight | Per capita disappearance, retail weight | (LBS) Pounds | 16.0712 | Green |
| Milk, Class III milk, All production practices, No Physical Attribute, Price received | Price received | $ / cwt | 16.55 | Green |
| Eggs and egg products, No Util Practice, No Prod Practice, No Physical Attribute, Imports | Imports | Million dozen,  shell-egg equivalent | 18 | Green |
| Lamb and mutton, No Util Practice, No Prod Practice, No Physical Attribute, Beginning stocks | Beginning stocks | Million LBS, carcass-weight equivalent | 32 | Green |
| Cattle,  Steers and heifers, All utilization practices, All production practices, No Physical Attribute, Feed price ratio | Feed price ratio | Bu corn to cwt live weight | 32.6 | Green |
| Lamb and mutton, No Util Practice, No Prod Practice, No Physical Attribute, Ending stocks | Ending stocks | Million LBS, carcass-weight equivalent | 33 | Green |
| Lamb and mutton, No Util Practice, No Prod Practice, No Physical Attribute, Beginning stocks | Beginning stocks | Million LBS, carcass-weight equivalent | 34.752 | Green |
| Milk, All utilization practices, All production practices, No Physical Attribute, Commercial exports,  skim-solid basis | Commercial exports,  skim-solid basis | Billion lbs, milk equivalent | 43.57 | Green |
| Milk, All utilization practices, All production practices, No Physical Attribute, Commercial exports, skim-solid basis | Commercial exports, skim-solid basis | Billion lbs, milk equivalent | 44.3839 | Green |
| Cattle, Cows, Slaughter, No Prod Practice, Physical Attributes Combination (Aggregate), Weighted average wholesale price | Weighted average wholesale price | $ / cwt, live basis | 54.5 | Green |
| Beef, No Util Practice, No Prod Practice, No Physical Attribute, Per capita disappearance, retail weight | Per capita disappearance, retail weight | (LBS) Pounds | 57.6771 | Green |
| Cattle, Cows, Slaughter, No Prod Practice, Physical Attributes Combination (Aggregate), Weighted average wholesale price | Weighted average wholesale price | $ / cwt, live basis | 59.595 | Green |
| Beef, No Util Practice, No Prod Practice, No Physical Attribute, Farm production | Farm production | Million LBS, carcass-weight equivalent | 65.7 | Green |
| Beef, No Util Practice, No Prod Practice, No Physical Attribute, Farm production | Farm production | Million LBS, carcass-weight equivalent | 66 | Green |
| Chickens, Other, No Util Practice, No Prod Practice, No Physical Attribute, Exports | Exports | Million LBS | 72 | Green |
| Chickens, Other, No Util Practice, No Prod Practice, No Physical Attribute, Exports | Exports | Million LBS | 76 | Green |
| Beef, No Util Practice, No Prod Practice, No Physical Attribute, Per capita disappearance, carcass weight | Per capita disappearance, carcass weight | (LBS) Pounds | 82.3959 | Green |
| Veal, No Util Practice, No Prod Practice, No Physical Attribute, Total disappearance | Total disappearance | Million LBS, carcass-weight equivalent | 85 | Green |
| Total red meat, No Util Practice, No Prod Practice, No Physical Attribute, Farm production | Farm production | Million LBS, carcass-weight equivalent | 89.9 | Green |
| Eggs and egg products, No Util Practice, No Prod Practice, No Physical Attribute, Beginning stocks | Beginning stocks | Million dozen, shell-egg equivalent | 92 | Green |
| Chickens,  Broilers, No Util Practice, No Prod Practice, No Physical Attribute, Per capita disappearance,  retail weight | Per capita disappearance, retail weight | (LBS) Pounds | 93.2381 | Green |
| Veal, No Util Practice, No Prod Practice, No Physical Attribute, Total supply | Total supply | Million LBS, carcass-weight equivalent | 94 | Green |
| Chickens,  Broilers, No Util Practice, No Prod Practice, No Physical Attribute, Per capita disappearance,  retail weight | Per capita disappearance, retail weight | (LBS) Pounds | 94.0508 | Green |
| Eggs and egg products, No Util Practice, No Prod Practice, No Physical Attribute, Ending stocks | Ending stocks | Million dozen, shell-egg equivalent | 95 | Green |
| Eggs,  Table, Wholesale market, Delivered retail store, Physical Attributes Combination (Aggregate), Wholesale price | Wholesale price | Cents Per Dozen | 105 | Green |
| Chickens,  Broilers, No Util Practice, No Prod Practice, No Physical Attribute, Per capita disappearance,  carcass weight | Per capita disappearance, carcass weight | (LBS) Pounds | 108.5426 | Green |
| Eggs and egg products, No Util Practice, No Prod Practice, No Physical Attribute, Beginning stocks | Beginning stocks | Million dozen, shell-egg equivalent | 108.7 | Green |
| Chickens,  Broilers, No Util Practice, No Prod Practice, No Physical Attribute, Per capita disappearance,  carcass weight | Per capita disappearance, carcass weight | (LBS) Pounds | 109.4887 | Green |
| Total poultry, No Util Practice, No Prod Practice, No Physical Attribute, Per capita disappearance,  retail weight | Per capita disappearance, retail weight | (LBS) Pounds | 110.7487 | Green |
| Total poultry, No Util Practice, No Prod Practice, No Physical Attribute, Per capita disappearance,  retail weight | Per capita disappearance, retail weight | (LBS) Pounds | 111.3962 | Green |
| Total red meat, No Util Practice, No Prod Practice, No Physical Attribute, Per | Per capita disappearance, | (LBS) Pounds | 111.8659 | Green |
| capita disappearance, retail weight | retail weight | | | |
| Cattle, Steers, Slaughter, Feedlot, negotiated purchase, Grade, Weighted average wholesale price | Weighted average wholesale price | $ / cwt, live basis | 120.75 | Green |
| Total poultry, No Util Practice, No Prod Practice, No Physical Attribute, Per capita disappearance, carcass weight | Per capita disappearance, carcass weight | (LBS) Pounds | 126.0532 | Green |
| Total poultry, No Util Practice, No Prod Practice, No Physical Attribute, Per capita disappearance, carcass weight | Per capita disappearance, carcass weight | (LBS) Pounds | 126.8341 | Green |
| Eggs, Table, Wholesale market, Delivered retail store, Physical Attributes Combination (Aggregate), Wholesale price | Wholesale price | Cents Per Dozen | 129.5125 | Green |
| Chickens, Broilers, No Util Practice, No Prod Practice, No Physical Attribute, Imports | Imports | Million LBS | 132 | Green |
| Sheep, Lambs, Slaughter, No Prod Practice, Physical Attributes Combination (Aggregate), Weighted average wholesale price | Weighted average wholesale price | $ / cwt, live basis | 145.75 | Green |
| Lamb and mutton, No Util Practice, No Prod Practice, No Physical Attribute, Commercial production | Commercial production | Million LBS, carcass-weight equivalent | 148 | Green |
| Total poultry, No Util Practice, No Prod Practice, No Physical Attribute, Imports | Imports | Million LBS | 151 | Green |
| Total red meat, No Util Practice, No Prod Practice, No Physical Attribute, Per capita disappearance, carcass weight | Per capita disappearance, carcass weight | (LBS) Pounds | 152.0244 | Green |
| Lamb and mutton, No Util Practice, No Prod Practice, No Physical Attribute, Total production | Total production | Million LBS, carcass-weight equivalent | 153.2 | Green |
| Total poultry, No Util Practice, No Prod Practice, No Physical Attribute, Imports | Imports | Million LBS | 155.668 | Green |
| Milk, All utilization practices, All production practices, No Physical Attribute, Domestic commercial use, | Domestic commercial use, skim- | Billion lbs, milk equivalent | 181.4578 | Green |

| skim-solid basis | solid basis | | | |
|---|---|---|---|---|
| Milk, All utilization practices, All production practices, No Physical Attribute, Production | Production | Billion lbs | 222.362 | Green |
| Milk, All utilization practices, All production practices, No Physical Attribute, Production | Production | Billion lbs | 222.7269 | Green |
| Turkeys, No Util Practice, No Prod Practice, No Physical Attribute, Ending stocks | Ending stocks | Million LBS | 235 | Green |
| Lamb and mutton, No Util Practice, No Prod Practice, No Physical Attribute, Imports | Imports | Million LBS, carcass-weight equivalent | 271 | Green |
| Eggs and egg products, No Util Practice, No Prod Practice, No Physical Attribute, Per capita disappearance | Per capita disappearance | Number, shell-egg equivalent | 292.758 | Green |
| Turkeys, No Util Practice, No Prod Practice, No Physical Attribute, Beginning stocks | Beginning stocks | Million LBS | 305 | Green |
| Turkeys, No Util Practice, No Prod Practice, No Physical Attribute, Ending stocks | Ending stocks | Million LBS | 305 | Green |
| Lamb and mutton, No Util Practice, No Prod Practice, No Physical Attribute, Total disappearance | Total disappearance | Million LBS, carcass-weight equivalent | 410.752 | Green |
| Lamb and mutton, No Util Practice, No Prod Practice, No Physical Attribute, Total disappearance | Total disappearance | Million LBS, carcass-weight equivalent | 416.2 | Green |
| Lamb and mutton, No Util Practice, No Prod Practice, No Physical Attribute, Total supply | Total supply | Million LBS, carcass-weight equivalent | 456.2 | Green |
| Lamb and mutton, No Util Practice, No Prod Practice, No Physical Attribute, Total supply | Total supply | Million LBS, carcass-weight equivalent | 456.652 | Green |
| Chickens, Broilers, No Util Practice, No Prod Practice, No Physical Attribute, Condemnations | Condemnations | Million LBS | 465.985 | Green |
| Total poultry, No Util Practice, No Prod Practice, No Physical Attribute, Condemnations | Condemnations | Million LBS | 466.5692 | Green |
| Chickens, Broilers, No Util Practice, No Prod Practice, No Physical Attribute, Condemnations | Condemnations | Million LBS | 468.4246 | Green |
| Total poultry, No Util Practice, No Prod Practice, No Physical Attribute, Condemnations | Condemnations | Million LBS | 469.0102 | Green |
| Chickens, Other, No Util Practice, No Prod Practice, No Physical Attribute, Total disappearance | Total disappearance | Million LBS | 471.776 | Green |
| Chickens, Other, No Util Practice, No Prod Practice, No Physical Attribute, Total disappearance | Total disappearance | Million LBS | 477.0144 | Green |
| Pork, No Util Practice, No Prod Practice, No Physical Attribute, Beginning stocks | Beginning stocks | Million LBS, carcass-weight equivalent | 540 | Green |
| Pork, No Util Practice, No Prod Practice, No Physical Attribute, Ending stocks | Ending stocks | Million LBS, carcass-weight equivalent | 555 | Green |
| Pork, No Util Practice, No Prod Practice, No Physical Attribute, Ending stocks | Ending stocks | Million LBS, carcass-weight equivalent | 580 | Green |
| Turkeys, No Util Practice, No Prod Practice, No Physical Attribute, Exports | Exports | Million LBS | 630 | Green |
| Beef, No Util Practice, No Prod Practice, No Physical Attribute, Ending stocks | Ending stocks | Million LBS, carcass-weight equivalent | 660 | Green |
| Beef, No Util Practice, No Prod Practice, No Physical Attribute, Beginning stocks | Beginning stocks | Million LBS, carcass-weight equivalent | 675 | Green |
| Beef, No Util Practice, No Prod Practice, No Physical Attribute, Ending stocks | Ending stocks | Million LBS, carcass-weight equivalent | 675 | Green |
| Chickens,  Broilers, No Util Practice, No Prod Practice, No Physical Attribute, Beginning stocks | Beginning stocks | Million LBS | 845 | Green |
| Pork, No Util Practice, No Prod Practice, No Physical Attribute, Imports | Imports | Million LBS, carcass-weight equivalent | 965 | Green |
| Eggs and egg products, No Util Practice, No Prod Practice, No Physical Attribute, Hatching use | Hatching use | Million dozen,  shell-egg equivalent | 1078.9 | Green |
| Eggs and egg products, No Util Practice, No Prod Practice, No Physical Attribute, Hatching use | Hatching use | Million dozen,  shell-egg equivalent | 1090 | Green |
| Total poultry, No Util Practice, No Prod Practice, No Physical Attribute, Beginning stocks | Beginning stocks | Million LBS | 1155.615 | Green |
| Eggs and egg products, No Util Practice, No Prod Practice, No Physical Attribute, Hatching production | Hatching production | Million dozen,  shell-egg equivalent | 1170 | Green |
| Total red meat, No Util Practice, No Prod Practice, No Physical Attribute, Beginning stocks | Beginning stocks | Million LBS, carcass-weight equivalent | 1256 | Green |
| Total red meat, No Util Practice, No Prod Practice, No Physical Attribute, Ending stocks | Ending stocks | Million LBS, carcass-weight equivalent | 1257 | Green |
| Total red meat, No Util Practice, No Prod Practice, No Physical Attribute, Ending stocks | Ending stocks | Million LBS, carcass-weight equivalent | 1300 | Green |
| Eggs and egg products, No Util Practice, No Prod Practice, No Physical Attribute, | Federally inspected eggs | Million dozen,  shell- | 2356.1 | Green |
| Federally inspected eggs broken | broken | egg equivalent | | |
| Beef, No Util Practice, No Prod Practice, No Physical Attribute, Exports | Exports | Million LBS, carcass-weight equivalent | 2894.2 | Green |
| Beef, No Util Practice, No Prod Practice, No Physical Attribute, Exports | Exports | Million LBS, carcass-weight equivalent | 3245 | Green |
| Hogs, All utilization practices, No Prod Practice, Condition, Imports | Imports | 1000 head | 5240 | Green |
| Turkeys, No Util Practice, No Prod Practice, No Physical Attribute, Total disappearance | Total disappearance | Million LBS | 5326 | Green |
| Turkeys, No Util Practice, No Prod Practice, No Physical Attribute, Federally inspected production | Federally inspected production | Million LBS | 5833.7 | Green |
| Turkeys, No Util Practice, No Prod Practice, No Physical Attribute, Net ready-to-cook production | Net ready-to-cook production | Million LBS | 5833.7 | Green |
| Turkeys, No Util Practice, No Prod Practice, No Physical Attribute, Federally inspected production | Federally inspected production | Million LBS | 5940 | Green |
| Turkeys, No Util Practice, No Prod Practice, No Physical Attribute, Net ready-to-cook production | Net ready-to-cook production | Million LBS | 5940 | Green |
| Turkeys, No Util Practice, No Prod Practice, No Physical Attribute, Total supply | Total supply | Million LBS | 6261 | Green |
| Cattle, Cows, Milk, All utilization practices, All production practices, No Physical Attribute, Inventory | Inventory | 1000 head | 9355 | Green |
| Cattle, Cows, Milk, All utilization practices, All production practices, No Physical Attribute, Inventory | Inventory | 1000 head | 9365 | Green |
| Eggs and egg products, No Util Practice, No Prod Practice, No Physical Attribute, Total production | Total production | Million dozen, shell-egg | 9440 | Green |

| | | equivalent | | |
|---|---|---|---|---|
| Pork, No Util Practice, No Prod Practice, No Physical Attribute, Total disappearance | Total disappearance | Million LBS, carcass-weight equivalent | 21184.7 | Green |
| Pork, No Util Practice, No Prod Practice, No Physical Attribute, Commercial production | Commercial production | Million LBS, carcass-weight equivalent | 27436 | Green |
| Pork, No Util Practice, No Prod Practice, No Physical Attribute, Total production | Total production | Million LBS, carcass-weight equivalent | 27450.2 | Green |
| Pork, No Util Practice, No Prod Practice, No Physical Attribute, Total supply | Total supply | Million LBS, carcass-weight equivalent | 28912.7 | Green |
| Chickens, Broilers, No Util Practice, No Prod Practice, No Physical Attribute, Total disappearance | Total disappearance | Million LBS | 35971.02 | Green |
| Chickens, Broilers, No Util Practice, No Prod Practice, No Physical Attribute, Total disappearance | Total disappearance | Million LBS | 36183.26 | Green |
| Total poultry, No Util Practice, No Prod Practice, No Physical Attribute, Total disappearance | Total disappearance | Million LBS | 41774.03 | Green |
| Total poultry, No Util Practice, No Prod Practice, No Physical Attribute, Total disappearance | Total disappearance | Million LBS | 41915.49 | Green |
| Chickens, Broilers, No Util Practice, No Prod Practice, No Physical Attribute, Net ready-to-cook production | Net ready-to-cook production | Million LBS | 43084.02 | Green |
| Chickens, Broilers, No Util Practice, No Prod Practice, No Physical Attribute, Net ready-to-cook production | Net ready-to-cook production | Million LBS | 43309.58 | Green |
| Chickens, Broilers, No Util Practice, No Prod Practice, No Physical Attribute, Federally inspected production | Federally inspected production | Million LBS | 43550 | Green |
| Chickens, Broilers, No Util Practice, No | Federally | Million LBS | 43778 | Green |
| Prod Practice, No Physical Attribute, Federally inspected production | inspected production | | | |
| Chickens,  Broilers, No Util Practice, No Prod Practice, No Physical Attribute, Total supply | Total supply | Million LBS | 44061.02 | Green |
| Chickens,  Broilers, No Util Practice, No Prod Practice, No Physical Attribute, Total supply | Total supply | Million LBS | 44386.26 | Green |
| Total poultry, No Util Practice, No Prod Practice, No Physical Attribute, Net ready-to-cook production | Net ready-to-cook production | Million LBS | 49569.41 | Green |
| Total poultry, No Util Practice, No Prod Practice, No Physical Attribute, Net ready-to-cook production | Net ready-to-cook production | Million LBS | 49689.95 | Green |
| Total poultry, No Util Practice, No Prod Practice, No Physical Attribute, Federally inspected production | Federally inspected production | Million LBS | 50036 | Green |
| Total poultry, No Util Practice, No Prod Practice, No Physical Attribute, Federally inspected production | Federally inspected production | Million LBS | 50158.98 | Green |
| Total poultry, No Util Practice, No Prod Practice, No Physical Attribute, Total supply | Total supply | Million LBS | 50876.03 | Green |
| Total poultry, No Util Practice, No Prod Practice, No Physical Attribute, Total supply | Total supply | Million LBS | 51020.39 | Green |
| Total red meat, No Util Practice, No Prod Practice, No Physical Attribute, Commercial production | Commercial production | Million LBS, carcass-weight equivalent | 53415.8 | Green |
| Total red meat, No Util Practice, No Prod Practice, No Physical Attribute, Total production | Total production | Million LBS, carcass-weight equivalent | 53505.7 | Green |
| Total red meat, No Util Practice, No Prod Practice, No Physical Attribute, Total supply | Total supply | Million LBS, carcass-weight equivalent | 58828.23 | Green |
| Chickens,  Other, No Util Practice, No | Condemnations | Million LBS | 0.5842 | Blue |
| Prod Practice, No Physical Attribute, Condemnations | | | | |
| Chickens, Other, No Util Practice, No Prod Practice, No Physical Attribute, Condemnations | Condemnations | Million LBS | 0.5856 | Blue |
| Eggs, All utilization practices, All production practices, No Physical Attribute, Feed price ratio | Feed price ratio | Lb laying feed to dozen eggs | 6.5 | Blue |
| Milk, All utilization practices, All production practices, No Physical Attribute, Ending commercial stocks, fat basis | Ending commercial stocks, fat basis | Billion lbs, milk equivalent | 13.1 | Blue |
| Veal, No Util Practice, No Prod Practice, No Physical Attribute, Commercial production | Commercial production | Million LBS, carcass-weight equivalent | 80 | Blue |
| Veal, No Util Practice, No Prod Practice, No Physical Attribute, Total production | Total production | Million LBS, carcass-weight equivalent | 85 | Blue |
| Cattle, Steers, Feeder, No Prod Practice, Physical Attributes Combination (Aggregate), Weighted average wholesale price | Weighted average wholesale price | $ / cwt, live basis | 149.75 | Blue |
| Eggs and egg products, No Util Practice, No Prod Practice, No Physical Attribute, Exports | Exports | Million dozen, shell-egg equivalent | 311 | Blue |
| Chickens, Other, No Util Practice, No Prod Practice, No Physical Attribute, Net ready-to-cook production | Net ready-to-cook production | Million LBS | 545.3994 | Blue |
| Chickens, Other, No Util Practice, No Prod Practice, No Physical Attribute, Federally inspected production | Federally inspected production | Million LBS | 546 | Blue |
| Chickens, Other, No Util Practice, No Prod Practice, No Physical Attribute, Net ready-to-cook production | Net ready-to-cook production | Million LBS | 546.675 | Blue |
| Chickens, Other, No Util Practice, No Prod Practice, No Physical Attribute, Federally inspected production | Federally inspected production | Million LBS | 547.277 | Blue |
| Chickens, Other, No Util Practice, No Prod Practice, No Physical Attribute, Total supply | Total supply | Million LBS | 554.0144 | Blue |
| Chickens, Other, No Util Practice, No Prod Practice, No Physical Attribute, Total supply | Total supply | Million LBS | 554.776 | Blue |
| Chickens, Broilers, No Util Practice, No Prod Practice, No Physical Attribute, Ending stocks | Ending stocks | Million LBS | 840 | Blue |
| Total poultry, No Util Practice, No Prod Practice, No Physical Attribute, Ending stocks | Ending stocks | Million LBS | 1082 | Blue |
| Total poultry, No Util Practice, No Prod Practice, No Physical Attribute, Ending stocks | Ending stocks | Million LBS | 1150 | Blue |
| Beef, No Util Practice, No Prod Practice, No Physical Attribute, Commercial production | Commercial production | Million LBS, carcass-weight equivalent | 25764 | Blue |
| Beef, No Util Practice, No Prod Practice, No Physical Attribute, Total disappearance | Total disappearance | Million LBS, carcass-weight equivalent | 25806.41 | Blue |
| Beef, No Util Practice, No Prod Practice, No Physical Attribute, Total production | Total production | Million LBS, carcass-weight equivalent | 25829.7 | Blue |
| Beef, No Util Practice, No Prod Practice, No Physical Attribute, Total supply | Total supply | Million LBS, carcass-weight equivalent | 29375.61 | Blue |
| Total red meat, No Util Practice, No Prod Practice, No Physical Attribute, Total disappearance | Total disappearance | Million LBS, carcass-weight equivalent | 47478.13 | Blue |